\documentclass[10pt,twocolumn,letterpaper]{article}

\usepackage[pagenumbers]{cvpr} 
\usepackage{multirow}
\usepackage{tabularx}
\usepackage{pifont}
\usepackage{bbm}
\usepackage[misc]{ifsym}

\definecolor{blue}{RGB}{0,0,255}
\definecolor{orange}{RGB}{249, 169, 104}
\definecolor{yellow}{RGB}{255, 255, 50}
\definecolor{reviewer1}{RGB}{246, 173, 105}
\definecolor{reviewer2}{RGB}{137, 181, 252}
\definecolor{reviewer3}{RGB}{127, 189, 120}

\newcommand{\teaser}{
\centering
\vspace{-7mm}
\large\url{https://neoverse-4d.github.io}\par
\vspace{5mm}
\includegraphics[width=1.0\textwidth,trim=0em 0em 0em 0em,clip]{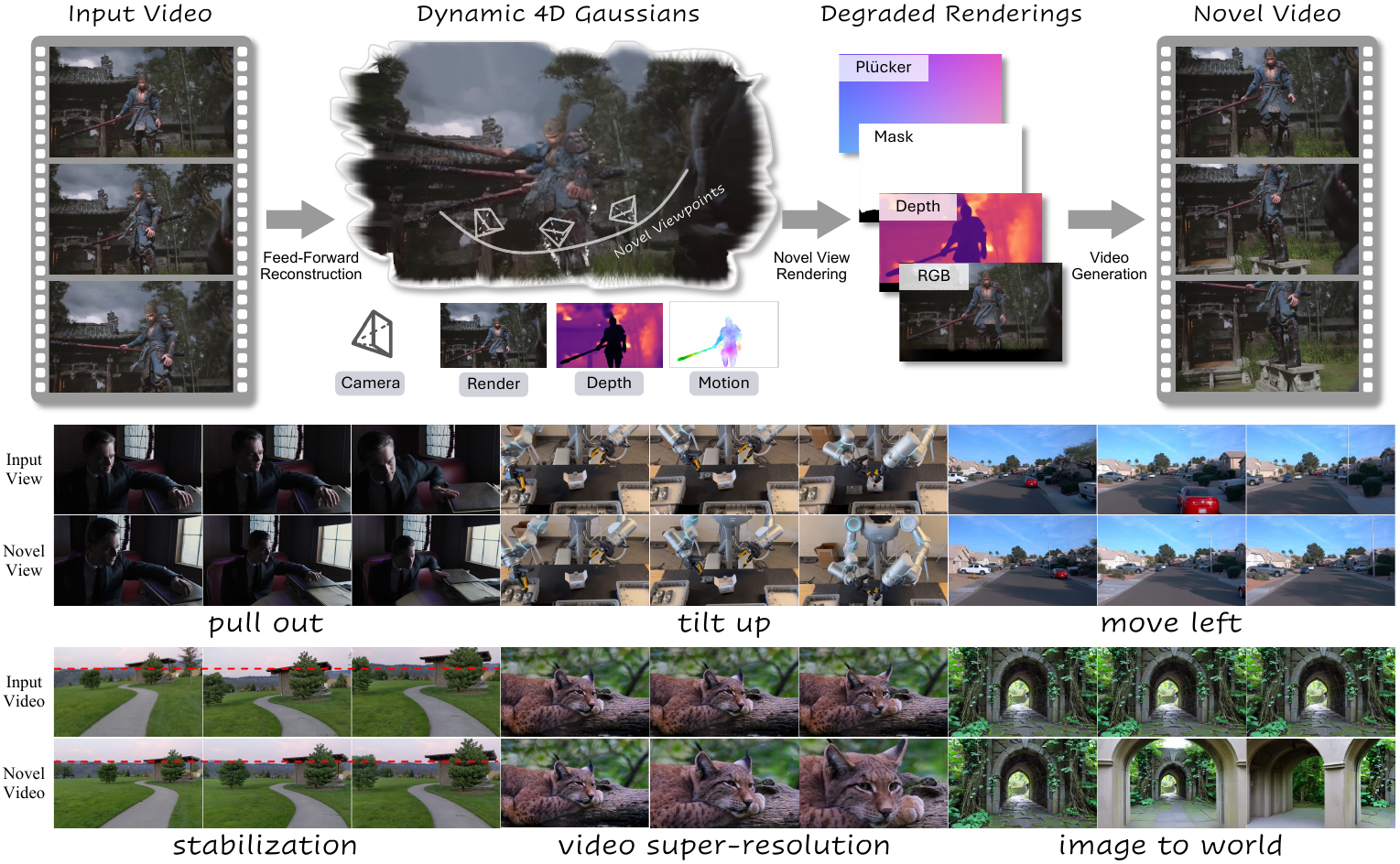}
\vspace{-1.2em}
\captionof{figure}{\textbf{Illustration of NeoVerse.}
NeoVerse reconstructs 4D Gaussian Splatting (4DGS) from monocular videos in a feed-forward manner.
These 4DGS can be rendered from novel viewpoints to provide degraded rendering conditions for generating high-quality and spatial-temporally coherent videos.
}
\vspace{0.7em}
\label{fig:teaser}
}

\definecolor{cvprblue}{rgb}{0.21,0.49,0.74}
\usepackage[pagebackref,breaklinks,colorlinks,allcolors=cvprblue]{hyperref}

\def\paperID{} 
\def\confName{CVPR}
\def\confYear{2026}

\def\name{NeoVerse }
\def\namenospace{NeoVerse}
\title{NeoVerse: Enhancing 4D World Model with in-the-wild Monocular Videos}

\author{%
\textbf{
  Yuxue Yang\textsuperscript{1,\,2}~~~~~Lue Fan\textsuperscript{1\,\Letter\,\,\dag}~~~~~Ziqi Shi\textsuperscript{1}~~~~~Junran Peng\textsuperscript{1}~~~~~Feng Wang\textsuperscript{2}~~~~~Zhaoxiang Zhang\textsuperscript{1\,\Letter}\vspace{1.5mm}}\\
  \textsuperscript{1}NLPR \& MAIS, CASIA~~~~~\textsuperscript{2}CreateAI\\
  \texttt{{\small\{yangyuxue2023, lue.fan\}@ia.ac.cn}}\\
  \vspace{-20pt}
}

\begin{document}
\twocolumn[
\maketitle
\teaser
]
{\let\thefootnote\relax\footnote{{\Letter} Corresponding Authors. \dag~ Project Lead.}}
\begin{abstract}

In this paper, we propose \textbf{\namenospace}, a versatile 4D world model that is capable of 4D reconstruction, novel-trajectory video generation, and rich downstream applications.
We first identify a common limitation of scalability in current 4D world modeling methods, caused either by expensive and specialized multi-view 4D data or by cumbersome training pre-processing.
In contrast, our \name is built upon a core philosophy that makes the full pipeline scalable to diverse in-the-wild monocular videos.
Specifically, \name features pose-free feed-forward 4D reconstruction, online monocular degradation pattern simulation, and other well-aligned techniques.
These designs empower \name with versatility and generalization to various domains.
Meanwhile, \name achieves state-of-the-art performance in standard reconstruction and generation benchmarks.
\end{abstract}
\section{Introduction}
\label{sec:intro}
4D world modeling holds transformative potential in many fields, such as digital content creation, autonomous driving, and embodied intelligence.
Recent approaches have made strides from both 3D side~\cite{gen3c,difix3d,viewcrafter,voyager,see3d, flexworld} and 4D side~\cite{trajectorycrafter,ex4d,uni3c,free4d,freesim,flexdrive,vivid4d, chronosobserver, ma2025follow, worldforge, postcam, van2024generative, cinemaster} with a principle of \emph{hybrid reconstruction and generation}.
This paradigm typically involves two stages: reconstructing a 3D/4D representation~\cite{depthcrafter,yang2024depth,bochkovskii2024depth,3dgs,monst3r} of the scene, and then, using the geometric prior to guide generation models~\cite{stable_diffusion,cogvideo,cogvideox,hunyuan_dit,wan}.
Such a reconstruction-generation hybrid paradigm has widely recognized promising features, including spatiotemporal consistency and precise viewpoint control. 
However, the current solutions usually have limitations in terms of \emph{\textbf{scalability}}.

The limitation of scalability manifests in two main aspects.
(1) \textbf{\emph{Limited data scalability}}.
Some methods, such as ViewCrafter~\cite{viewcrafter}, utilize videos of static scenes to create multi-view training data and learn to generate videos in novel trajectories.
Although effective, they cannot be extended to 4D scenes.
Some other methods, such as SynCamMaster~\cite{syncammaster}, CamCloneMaster~\cite{camclonemaster}, and ReCamMaster~\cite{recammaster}, depend on specialized, hard-to-capture multi-view dynamic videos to learn novel trajectory generation.
Such non-scalable data limits the model's generalization and versatility.
(2) \textbf{\emph{Limited training scalability}}.
Another line of work~\cite{trajectorycrafter, freesim,uni3c,vivid4d} utilizes more flexible data types but usually necessitates a cumbersome offline pre-processing stage to create training data.
For example, TrajectoryCrafter generates training data using a heavy video depth estimator~\cite{depthcrafter} in an offline manner.
Similarly, previous work FreeSim~\cite{freesim} pre-reconstructs the Gaussian field to prepare training input, which utilizes offline reconstruction~\cite{pvg, omnire} and may even rely on extra 3D detection methods~\cite{centerpoint, bevformer, sst, mixsup}.
Such an offline curation usually leads to significant computational burden, storage consumption, inflexible training scheme tuning, and even disables online augmentations.
The two kinds of limitations erect a barrier to leveraging the cheap and diverse in-the-wild monocular videos, constraining the potential for building more powerful models.

To address these challenges, we propose \namenospace.
The core philosophy of \name is \emph{\textbf{making the full pipeline scalable to diverse in-the-wild monocular videos}}, enhancing generalization and versatility of 4D world models.
To implement our vision, we first propose a feed-forward 4DGS model, built upon VGGT~\cite{vggt}.
This model not only ``Gaussianizes'' VGGT but also features a bidirectional motion modeling mechanism, which is crucial for efficient online reconstruction (Sec.~\ref{sec:gen}) and applications requiring time control.
We then incorporate this feed-forward model into the generation training process.
During each training iteration, it efficiently reconstructs 4D scenes using sparse key frames from monocular videos in an online manner.
In addition, efficient online monocular degradation simulations, including Gaussian culling and average geometry filter, are proposed to simulate degraded rendering patterns in novel trajectories and offer conditions for generation.
Combining them together makes the whole training process scalable to diverse in-the-wild monocular videos (up to 1M clips) in terms of both training efficiency and technical feasibility.
We summarize our contributions as follows.
\begin{itemize}
\item We propose \namenospace, a 4D world modeling approach, which is scalable to and enhanced by diverse in-the-wild monocular videos.
\item \name is versatile, enabling many applications, including 4D reconstruction, multiview video generation, video editing, stabilization, super-resolution, etc.
\item \name achieves state-of-the-art results in both reconstruction and generation tasks. 
\item We will make the source code publicly available to decentralize general 4D world models by leveraging cheap and diverse in-the-wild monocular videos.
\end{itemize}
\section{Related Works}
\paragraph{Feed-forward Gaussian reconstruction.}
Recent stereo and 3D geometry foundation models~\cite{dust3r, noposplat, monst3r, vggt, anysplat, mapanything, worldmirror, da3, 4dgt, streamsplat, movies} can estimate dense depth, point maps, and even camera parameters in a single forward pass, thereby driving a shift in Gaussian Splatting from per-scene optimization to generalizable feed-forward reconstruction.
For static scenes, pose-free models such as NoPoSplat~\cite{noposplat} reconstruct 3D Gaussians directly from sparse, unposed multi-view images, and AnySplat~\cite{anysplat} further extends this paradigm to casually captured, long uncalibrated image sequences. For dynamic scenes, 4DGT~\cite{4dgt}, StreamSplat~\cite{streamsplat}, and MoVieS~\cite{movies} push feed-forward GS into 4D; however, each method still retains specific constraints: 4DGT is trained on posed monocular videos and adopts a largely uni-directional temporal modeling strategy, MoVieS similarly assumes known camera poses during training and inference, while StreamSplat focuses on frame-by-frame modeling.

\paragraph{Reconstruction-based video generation.}
Recent methods such as GEN3C~\cite{gen3c}, DaS~\cite{das}, See3D~\cite{see3d}, ViewCrafter~\cite{viewcrafter}, Difix3D+~\cite{difix3d}, GS-DiT~\cite{gsdit}, Voyager~\cite{voyager}, Uni3C~\cite{uni3c}, FreeSim~\cite{freesim}, TrajectoryCrafter~\cite{trajectorycrafter}, See4D~\cite{see4d}, PostCam~\cite{postcam}, Light-X~\cite{lightx} follow a hybrid \emph{reconstruction+generation} paradigm, where a 3D/4D representation is first reconstructed and then used as geometric guidance for a generative video model.
GEN3C~\cite{gen3c} builds a depth-based 3D feature cache whose renderings condition a video diffusion model for 3D-consistent, pose-controllable synthesis; ViewCrafter~\cite{viewcrafter} adopts a point-conditioned video diffusion framework to extend single- or sparse-view inputs into long-range, high-fidelity novel-view sequences; Difix3D+~\cite{difix3d} applies a single-step diffusion enhancer to rendered novel views to correct artifacts in underconstrained regions and distill the improvements back into NeRF/3DGS representations; and TrajectoryCrafter~\cite{trajectorycrafter} formulates camera-controllable video generation for monocular videos as trajectory redirection, conditioning a dual-stream diffusion backbone on point-cloud renderings and source frames to follow user-specified camera paths. Despite their strong spatial–temporal consistency and viewpoint controllability, these reconstruction-based approaches are mostly tailored to static or quasi-static scenes and rely on curated data or heavyweight offline reconstruction, limiting scalability to in-the-wild monocular videos.

\section{Methodology}
This section is organized as follows. In Sec.~\ref{sec:recon}, we first propose an efficient pose-free feed-forward 4DGS reconstruction model, which reconstructs 4DGS from monocular videos. In Sec.~\ref{sec:gen}, we introduce how to combine reconstruction part and generation and make the full pipeline scalable.
Sec.~\ref{sec:train} contains the training scheme
and Sec.~\ref{sec:infer} elaborates on inference strategies.
\begin{figure*}[ht]
    \centering
    \includegraphics[width=\linewidth]{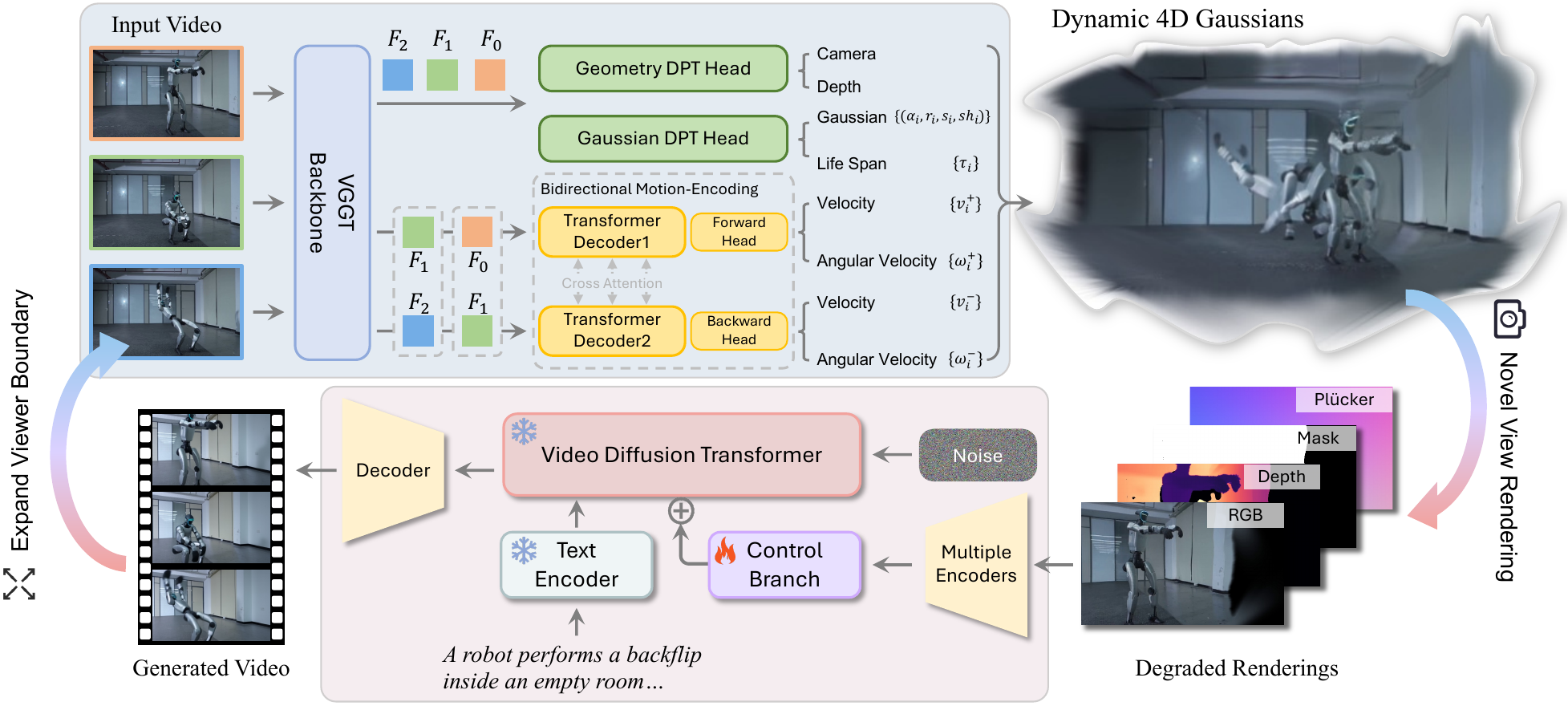}
    \vspace{-5mm}
    \caption{\textbf{Framework of \namenospace.} In the reconstruction part, we propose a pose-free feed-forward 4DGS reconstruction model (\cref{sec:recon}) with bidirectional motion modeling.
    The degraded renderings in novel viewpoints from 4DGS are input to the generation model as conditions.
    During training, the degraded rendering conditions are simulated from monocular videos (\cref{sec:gen}), and the original videos themselves serve as targets.}
    \vspace{-5mm}
    \label{fig:framework}
\end{figure*}
\subsection{Pose-Free Feed-Forward 4DGS Reconstruction}
\label{sec:recon}
Our feed-forward model is partially built upon VGGT~\cite{vggt} backbone.
For simplicity, we mainly introduce how we make VGGT dynamic and ``Gaussianized''.
\vspace{-3mm}
\paragraph{Bidirectional motion modeling.}
Given a monocular video $\{\boldsymbol{I}_t\in\mathbb{R}^{H\times W\times 3}\}_{t=1}^T$, VGGT extracts the frame-wise features using the pretrained DINOv2~\cite{dinov2}.
These features, concatenated with camera tokens and register tokens, are fed into a series of Alternating-Attention blocks~\cite{vggt}, obtaining so-called \emph{\textbf{frame features}}.
While this process effectively aggregates spatial information, they are insufficient for motion modeling due to temporal unawareness.

We introduce a bidirectional motion-encoding branch.
Different from uni-directional motion in 4DGT~\cite{4dgt}, the bidirectional prediction distinguishes the instantaneous velocity between $t \rightarrow t+1$ and $t \rightarrow t-1$.
Such a distinction facilitates temporal Gaussian interpolation between two consecutive timestamps.

Specifically, for the frame features $\{\boldsymbol{F}_t\}_{t=1}^T$, we copy and slice them into two parts along the temporal dimension: $\{\boldsymbol{F}_t\}_{t=1}^{T-1}$ and $\{\boldsymbol{F}_t\}_{t=2}^T$.
Then we obtain \textbf{forward motion features} using the first part as queries and the second part as keys and values.
Similarly, the \textbf{backward motion features} are encoded conversely.
Formally, we have
\begin{equation}
\begin{aligned}
\{\boldsymbol{F}_t^\text{fwd}\}_{t=1}^{T-1} &= \text{CrossAttn}(q=\{\boldsymbol{F}_t\}_{t=1}^{T-1}; k,v=\{\boldsymbol{F}_t\}_{t=2}^{T}), \\
\{\boldsymbol{F}_t^\text{bwd}\}_{t=2}^{T} &= \text{CrossAttn}(q=\{\boldsymbol{F}_t\}_{t=2}^{T}; k,v=\{\boldsymbol{F}_t\}_{t=1}^{T-1}),
\end{aligned}
\label{eq:motion}
\end{equation}
where $\boldsymbol{F}_t^\text{fwd}$ and $\boldsymbol{F}_t^\text{bwd}$ are forward motion features from timestamp $t$ to $t+1$, and backward motion features from $t$ to $t-1$.
These features will be utilized to predict bidirectional linear and angular velocity of Gaussian primitives.

\paragraph{Gaussianizing VGGT.}
We first define 4D Gaussians as
\begin{equation}
\{(\boldsymbol{\mu}_i, \boldsymbol{\alpha}_i, \boldsymbol{r}_i, \boldsymbol{s}_i, \boldsymbol{sh}_i, \boldsymbol{\tau}_i, \boldsymbol{v}^+_i, \boldsymbol{v}^-_i, \boldsymbol{\omega}^+_i, \boldsymbol{\omega}^-_i)\}_{i=1}^{T\times H\times W},
\end{equation}
where each Gaussian $i$ is parameterized by: 3D position $\boldsymbol{\mu}_i$, opacity $\boldsymbol{\alpha}_i$, rotation $\boldsymbol{r}_i$, scale $\boldsymbol{s}_i$, and spherical harmonics coefficients $\boldsymbol{sh}_i$, as inherited from 3D Gaussians~\cite{3dgs}.
For bidirectional motion modeling, we introduce forward and backward velocities $\boldsymbol{v}^+_i$, $\boldsymbol{v}^-_i$, and forward and backward angular velocities $\boldsymbol{\omega}^+_i$, $\boldsymbol{\omega}^-_i$.
In addition, we adopt a life span $\boldsymbol{\tau}_i$ following the common practice in 4DGS.

The 3D positions $\{\boldsymbol{\mu}_i\}$ is obtained by back-projecting pixel depth to 3D space using predicted depth and camera parameters.
For the other attributes, $\{(\boldsymbol{\mu}_i, \boldsymbol{\alpha}_i, \boldsymbol{r}_i, \boldsymbol{s}_i, \boldsymbol{sh}_i, \boldsymbol{\tau}_i\}$ are predicted from the frame features, while the dynamic attributes $\{\boldsymbol{v}^+_i, \boldsymbol{v}^-_i, \boldsymbol{\omega}^+_i, \boldsymbol{\omega}^-_i\}$ are predicted from the bidirectional motion features.

\subsection{Reconstruction-Guided Video Generation}
\label{sec:gen}
In this subsection, we introduce how to combine the reconstruction and generation in a scalable training pipeline.
\vspace{-3mm}
\paragraph{Efficient on-the-fly reconstruction from sparse key frames.}
Although the proposed feed-forward 4DGS reconstruction is efficient, it can still be the bottleneck of training efficiency if we conduct on-the-fly reconstruction with long video input.
To boost the training efficiency, we propose reconstruction from sparse key frames.

Given a long video input with $N$ frames, we only take $K$ key frames as reconstruction input but \textbf{render from all the $N$ frames} since the rendering process is extremely efficient compared with network computation.
However, such an operation requires interpolating the Gaussian field at non-keyframes.
Thanks to our bidirectional motion modeling, such interpolation can be implemented as follows.

Given a non-key-frame query timestamp $t_q$, we transfer a nearest key-frame Gaussian $i$ at timestamp $t$ to $t_q$ following
\begin{align}
\vspace{-3mm}
\boldsymbol{\mu}_i(t_q) &=
    \begin{cases}
    \boldsymbol{\mu}_i + \boldsymbol{v}^+_i |t_q - t|, & t_q \geq t,\\
    \boldsymbol{\mu}_i + \boldsymbol{v}^-_i |t_q - t|, & t_q < t, \\
    \end{cases} \label{eq:mu}
    \vspace{-2mm}
\end{align}
\vspace{-3mm}
\begin{align}
\boldsymbol{r}_i(t_q) &=
    \begin{cases}
    \boldsymbol{r}_i \cdot \phi(\boldsymbol{\omega}^+_i |t_q - t|), & t_q \geq t,\\
    \boldsymbol{r}_i \cdot \phi(\boldsymbol{\omega}^-_i |t_q - t|), & t_q < t,
    \end{cases} \label{eq:r}
\end{align}
\vspace{-2mm}
\begin{align}
\boldsymbol{\alpha}_i(t_q) =\boldsymbol{\alpha}_i \text{exp}(-\gamma\cdot d(t_q, t)^\frac{1}{1-\boldsymbol{\tau_i}}),
\label{eq:alpha}
\vspace{-3mm}
\end{align}
where we assume the real-world motion in a short interval between two adjacent input frames is approximately linear.
Angular velocities $\boldsymbol{\omega}_i^\pm$ are represented in the axis-angle representation, and $\phi(\cdot)$ converts it to a quaternion.
The opacity of the Gaussian is represented by a time-varying function to ensure a natural transition between input frames.
To handle non-uniform keyframe intervals, we model opacity decay with a normalized temporal distance $d(t_q, t)=\frac{|t_q-t|}{|T_{k+1}-T_{k}|} \leq 1$, where $[T_k, T_{k+1}]$ is the keyframe interval containing query timestamp $t_q$.
The life span $\boldsymbol{\tau}_i$ is constrained in the range of $(0, 1)$ with a sigmoid function, and $\gamma$ is a hyper-parameter that controls the decay speed.
When $\boldsymbol{\tau}_i$ approaches 1, the $\text{exp}(\cdot)$ tends towards 1, indicating $\boldsymbol{\alpha}_i(t_q)\approx \boldsymbol{\alpha}_i$; otherwise, $\boldsymbol{\alpha}_i(t_q)$ decays rapidly.
\vspace{-3mm}
\paragraph{Monocular degradation simulation.}
\begin{figure}[t]
    \centering
    \includegraphics[width=\linewidth]{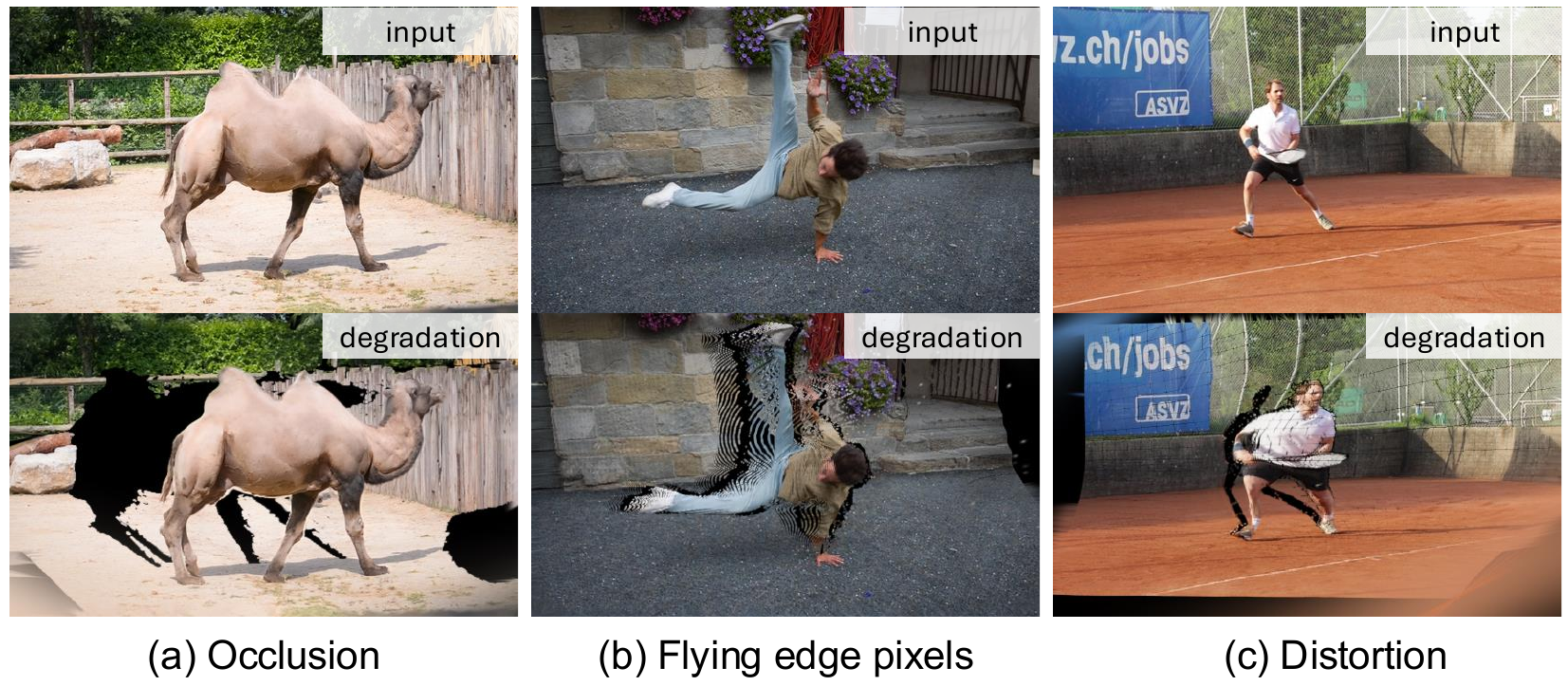}
    \caption{\textbf{Training pairs with degradation simulation.}
    }
    \vspace{-3mm}
    \label{fig:degradation}
\end{figure}
Our generation model is expected to generate high-quality novel views from low-quality novel view renderings, necessitating such training pairs.
For multi-view or static datasets~\cite{scannet++, dl3dv}, we can easily get such training pairs as in ViewCrafter~\cite{viewcrafter}.
However, for in-the-wild monocular videos, we need to carefully simulate degradation renderings paired with ground-truth monocular frames.
Therefore, we propose three techniques to simulate the degradation rendering patterns based on monocular videos.
\par
\emph{(1) \textbf{Visibility-based Gaussian Culling} for occlusion simulation.}
Given the camera pose trajectory predicted from the sparse key frames, we apply a random transform to the trajectory to obtain a novel trajectory.
A constraint is applied to this transform to ensure new camera poses still roughly point to the scene center.
Using depth, we can easily identify those Gaussians that are occluded from the transformed new camera poses.
We then simply cull those invisible Gaussian primitives and render the remaining Gaussian primitives back into the original viewpoints, resulting degradation pattern demonstrated in Fig.~\ref{fig:degradation} (a).

\emph{(2) \textbf{Average Geometry Filter} for flying-edge-pixel and distortion simulation.}
In addition to the occlusion, another typical degradation pattern is the flying pixels in depth-discontinuous edges.
The network has tendency to produce \emph{\textbf{average}} depth value at those edges to minimize regression loss, as also confirmed by~\cite{pixelperfect}.
From a first-principles perspective, we propose to use a \emph{\textbf{average filter}} to create such averaged depth patterns.
Specifically, we render depth in the transformed novel trajectory and apply an average filter in the rendered depth map.
We then adjust the center position of each Gaussian according to the average filtered depth value.
When such modified Gaussians are rendered back into the original views, the flying-pixel pattern appears as shown in Fig.~\ref{fig:degradation} (b).
We can further apply a larger filter kernel to simulate spatially broader distortions shown in Fig.~\ref{fig:degradation} (c),  caused by potential depth error.
\par
All three kinds of degradations in Fig.~\ref{fig:degradation} are simulated with fundamental principles in geometry relation and depth learning, and designed to be simple yet effective, enabling the utilization of in-the-wild monocular videos.

\paragraph{Degraded rendering conditioning.}
We use the obtained degraded renderings as conditions for generation and the original videos as targets.
The rendered conditions include multiple modalities, including RGB images, depth maps, and masks binarized from opacity maps to indicate the empty regions.
Plüker embeddings of the original trajectory are also computed to provide explicit 3D camera motion information~\cite{uni3c}.
We introduce a control branch to incorporate them into the generation model like~\cite{controlnet, vace, voyager, layeranimate}.
During training, we only train the control branch while freezing the video generation model, not only for training efficiency, but more importantly, to make \name accessible to powerful distillation LoRAs~\cite{lora} to speed up the generation process.

\subsection{Training Scheme}
\label{sec:train}
We partition the training into two stages: 1) reconstruction model training; 2) generation model training with on-the-fly reconstruction and degradation simulation.
\paragraph{Reconstruction.}
We train our feed-forward 4DGS reconstruction model with a multi-task loss on various static and dynamic 3D datasets:
\begin{equation}
\mathcal{L}_\text{recon} = \mathcal{L}_\text{rgb} + \lambda_\text{1}\mathcal{L}_\text{camera} + \lambda_\text{2}\mathcal{L}_\text{depth} + \lambda_\text{3}\mathcal{L}_\text{motion} +
\lambda_\text{4}\mathcal{L}_\text{regular},
\end{equation}
where $\mathcal{L}_\text{rgb}$ is the photometric loss between rendered and ground-truth images, including an $L_2$ loss and LPIPS~\cite{lpips} loss.
The camera loss $\mathcal{L}_\text{camera}$ and depth loss $\mathcal{L}_\text{depth}$ supervise the predicted camera parameters and depth maps following VGGT~\cite{vggt}.
Notably, $\mathcal{L}_\text{depth}$ also contains the supervision for rendered depth from Gaussians.
The motion loss $\mathcal{L}_\text{motion}=\sum_i \Vert\hat{\boldsymbol{v}}^+_i-\boldsymbol{v}^+_i\Vert + \Vert\hat{\boldsymbol{v}}^-_i-\boldsymbol{v}^-_i\Vert$ adds supervision on the predicted bidirectional velocities, where $\hat{\boldsymbol{v}}^+_i$ and $\hat{\boldsymbol{v}}^-_i$ are the ground-truth forward and backward velocities computed from some dynamic 3D datasets~\cite{pointodyssey, dynamicstereo, kubric, spring, waymo, vkitti2}.
To prevent the Gaussians from becoming erroneously transparent, we introduce a regularization loss $\mathcal{L}_\text{regular} = \sum_i \vert 1 - \boldsymbol{A}_i\vert$, where $\boldsymbol{A}_i$ is rendered accumulated opacity map.

\paragraph{Generation.}
For generation model training, we adopt Rectified Flow~\cite{rectified} and Wan-T2V~\cite{wan} 14B to model the denoising diffusion process.
\textbf{The whole training process is performed on monocular videos.}
Given a monocular video, we first utilize on-the-fly reconstruction from sparse key frames to obtain 4DGS and simulate degradation renderings as conditions $c_\text{render}$.
For the video latent $x_1$ and sampled noise $x_0\sim\mathcal{N}(0, I)$, the training objective of generation model $f_\theta$ is formulated as
\begin{equation}
\mathcal{L}_\text{gen} = \mathbb{E}_{x_1, x_0, c_\text{render}, c_\text{text}, t} \| f_\theta(x_t, t, c_\text{render}, c_\text{text}) - v_t\|_2^2 ,
\end{equation}
where $x_t$ is a linear interpolation between $x_1$ and $x_0$ at timestamp $t$, $v_t=x_1-x_0$ is ground-truth velocity.
$c_\text{text}$ is the text condition extracted from the video caption using a language model like umT5~\cite{umT5}.
Renderings $c_\text{render}$ are input into the generation model through a control branch like~\cite{controlnet, vace}.

\subsection{Inference}
\label{sec:infer}
\paragraph{Reconstruction and global motion tracking.}
Given a monocular video, our feed-forward model outputs 4DGS and camera parameters of each frame.
Before rendering conditions from a novel trajectory, we can optionally aggregate Gaussians from multiple timestamps into a single timestamp for a more complete representation.
For better aggregation, we conduct motion separation by global motion tracking.
\par
The motivation of global motion tracking is to identify those objects undergoing both static and dynamic phases in a clip, which should be regarded as the dynamic part and cannot be easily identified using predicted instantaneous velocity.
Taking a Gaussian primitive $i$ as example, given world-to-camera poses $\{\boldsymbol{P}_t\}_{t=1}^T$, camera intrinsics $\{\boldsymbol{K}_t\}_{t=1}^T$, and Gaussian position $\boldsymbol{\mu}_i$ for Gaussian $i$, we project the Gaussian center to each frame $t$ and compute its projected pixel coordinates $\boldsymbol{p}_{i,t}$ and depth $\boldsymbol{d}_{i,t}$.
Let $D_t[\boldsymbol{p}_{i,t}]$ and $V_t[\boldsymbol{p}_{i,t}]$ are the sampled depth and velocity at pixel $\boldsymbol{p}_{i,t}$.
We define a visibility-weighted maximum velocity magnitude at the global video level as
\begin{equation}
\vspace{-2mm}
\begin{aligned}
\boldsymbol{m}_{i,t} &= \max\{\|V_t^+[\boldsymbol{p}_{i,t}]\|_2, \|V_t^-[\boldsymbol{p}_{i,t}]\|_2\},\\
\boldsymbol{m}_i &= \max_{t=1,\dots,T} \mathbbm{1}(\boldsymbol{d}_{i,t} \le D_t[\boldsymbol{p}_{i,t}]) \cdot \boldsymbol{m}_{i,t},
\end{aligned}
\vspace{-2mm}
\end{equation}
where $\boldsymbol{m}_{i,t}$ is the maximum velocity magnitude at frame $t$, $\mathbbm{1}(\cdot)$ is a function indicating whether the Gaussian is visible, and $\boldsymbol{m}_i$ is the visibility-weighted maximum velocity magnitude across all frames.
Finally, we separate the Gaussians into static set $\mathcal{S}$ and dynamic set $\mathcal{D}$ according to $\boldsymbol{m}_i$ with a threshold $\eta$.

\paragraph{Temporal aggregation, interpolation, and generation.}
With a separated dynamic part and a static part, we conduct two different Gaussian temporal aggregation strategies for each part, respectively.
The static part is simply aggregated across all frames, while the dynamic part is aggregated only from a couple of nearby frames to avoid motion drifting errors.
\par
In some cases, we may need to interpolate Gaussians into an intermediate timestamp between two adjacent discrete frames.
A typical case is creating slow-motion videos and bullet-time shots.
Our bidirectional motion mechanism sufficiently supports such tasks happening in a short time interval.
In practice, we use similar techniques in Sec.~\ref{sec:gen} for interpolation.
\par
After the optional aggregation and interpolation, we render the resulting Gaussians into any desired novel trajectory.
The renderings, along with other conditions, are sent to the generation model to generate videos.

\section{Experiments}
\subsection{Implementation}
For reconstruction, we follow the learning rate schedule of VGGT~\cite{vggt}.
We resize all input videos to have a longest edge of 560 pixels.
GSplat~\cite{gsplat} is adopted as the Gaussian Splatting rendering backend.
For the generation, the video resolution is fixed at $336\times 560$ and the length is set to $81$ frames.
The training is conducted on 32 A800 GPUs, where the first stage trains 150K iterations and the second stage trains 50K iterations.
More training details can be found in the supplementary material.

\paragraph{Datasets.}
We collect 18 public datasets following CUT3R~\cite{cut3r}, including Arkitscenes~\cite{arkitscenes}, DL3DV~\cite{dl3dv}, PointOdyssey~\cite{pointodyssey}, Kubric~\cite{kubric}, Waymo~\cite{waymo}, SpatialVID~\cite{spatialvid}, GFIE~\cite{gfie}, etc.
Besides the above datasets, we further curate a large-scale self-collected monocular video dataset from the internet, containing over 1M videos from diverse scenarios.
More details about datasets are provided in the supplementary material.

\begin{figure*}[th]
    \centering
    \includegraphics[width=\linewidth]{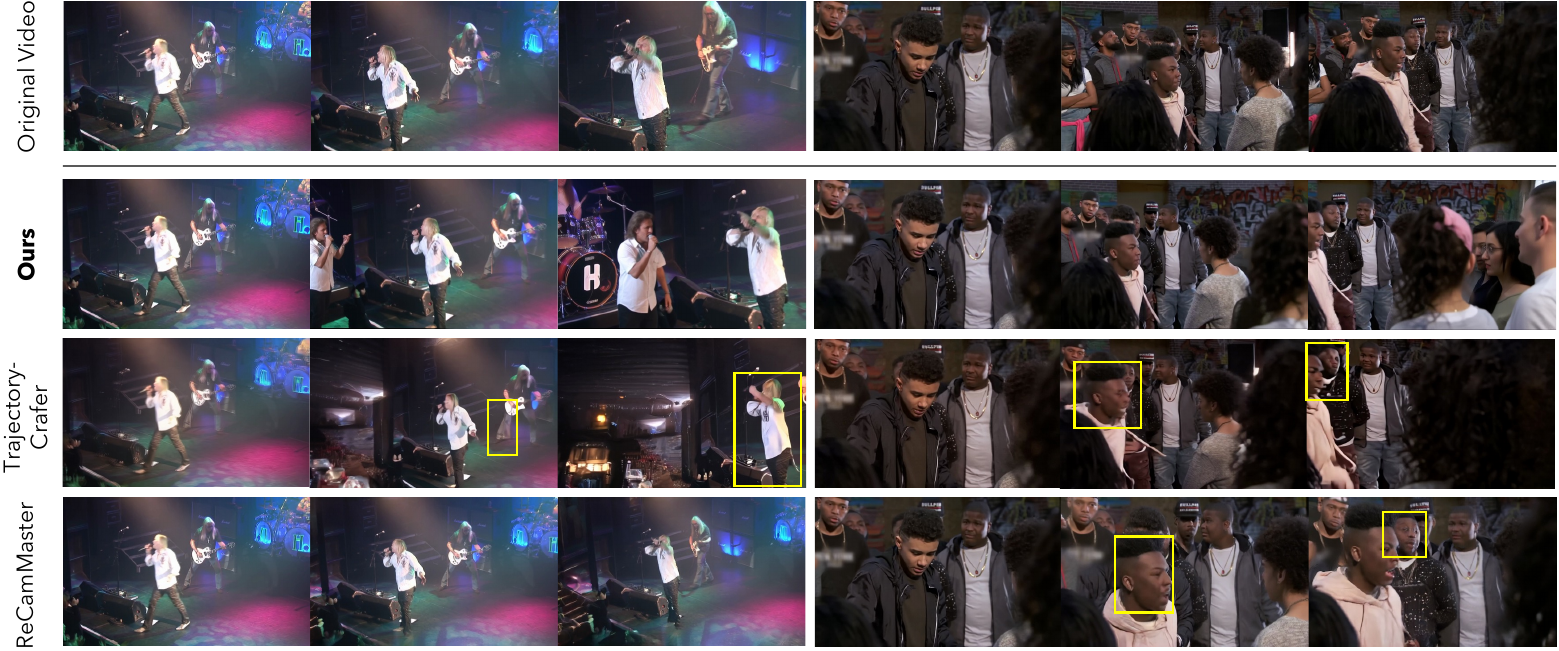}
    \vspace{-6mm}
    \caption{\textbf{Generation with large camera motions on challenging in-the-wild videos.} We compare our method against other related work on ``Pan left'' (left) and ``Move right'' (right) cases. Our \name achieves better generation quality while maintaining precise camera controllability. Yellow boxes highlight artifacts.}
    \vspace{-2mm}
    \label{fig:gen_comp}
\end{figure*}

\subsection{Quantitative Evaluation}
\paragraph{Reconstruction benchmark.}
Our reconstruction results on both static and dynamic datasets are shown in Table~\ref{tab:static_reconstruction} and Table~\ref{tab:dynamic_reconstruction}, respectively.
Our reconstruction part achieves state-of-the-art performance among all metrics. Recent reprints MoVieS~\cite{movies} and StreamSplat~\cite{streamsplat} are not listed in the table because they are neither open-sourced nor provide a detailed evaluation protocol.
Our detailed evaluation protocols are provided in the supplementary material.
\begin{table}[ht]
\vspace{5pt}
\centering
\footnotesize
\resizebox{\columnwidth}{!}{
\begin{tabular}{c|ccc|ccc}
\toprule
\multirow{2}{*}{Method} & \multicolumn{3}{c|}{VRNeRF~\cite{vrnerf} (16 views)}& \multicolumn{3}{c}{Scannet++~\cite{scannet++} (32 views)}\\
& PSNR$\uparrow$ & SSIM$\uparrow$ & LPIPS$\downarrow$ & PSNR$\uparrow$ & SSIM$\uparrow$ & LPIPS$\downarrow$\\
\midrule
NoPoSplat~\cite{noposplat} &11.27&0.408&0.620&8.69&0.312&0.614\\
Flare~\cite{flare} &12.62&0.597&0.623&12.19&0.619&0.611\\
AnySplat~\cite{anysplat} &18.02&0.705&0.366&22.79&0.773&0.217\\
\textbf{Ours} &\textbf{20.73}&\textbf{0.766}&\textbf{0.352}&\textbf{25.34}&\textbf{0.834}&\textbf{0.195}\\
\bottomrule
\end{tabular}
}
\caption{Quantitative comparison with other \textbf{static} reconstruction models.}
\vspace{-13pt}
\label{tab:static_reconstruction}
\end{table}

\begin{table}[h]
\vspace{5pt}
\centering
\footnotesize
\resizebox{\columnwidth}{!}{
\begin{tabular}{c|ccc|ccc}
\toprule
\multirow{2}{*}{Method} & \multicolumn{3}{c|}{ADT~\cite{adt}}& \multicolumn{3}{c}{DyCheck~\cite{dycheck}}\\
& PSNR$\uparrow$ & SSIM$\uparrow$ & LPIPS$\downarrow$ & PSNR$\uparrow$ & SSIM$\uparrow$ & LPIPS$\downarrow$\\
\midrule
MonST3R~\cite{monst3r} &17.42&0.554&0.534&9.32&0.103&0.710\\
4DGT$^\dagger$~\cite{4dgt} &30.09&0.909&0.178&9.94&0.208&0.639\\
\textbf{Ours} &\textbf{32.56}&\textbf{0.927}&\textbf{0.120}&\textbf{11.56}&\textbf{0.293}&\textbf{0.558}\\
\bottomrule
\end{tabular}
}
\caption{Quantitative comparison with other \textbf{dynamic} reconstruction models. $^\dagger$: indicate the method takes camera poses as input.}
\vspace{-13pt}
\label{tab:dynamic_reconstruction}
\end{table}

\paragraph{Generation benchmark.}
In Table~\ref{tab:generation}, we compare the generation performance with related work TrajectoryCrafter~\cite{trajectorycrafter} and ReCamMaster~\cite{recammaster}, demonstrating better performance.
We conduct more analysis in the section of qualitative evaluation.

\paragraph{Runtime evaluation.}
Table~\ref{tab:generation} also shows the efficiency evaluation of both the reconstruction stage and the generation stage.
Thanks to our intentional design of condition injection in Sec.~\ref{sec:gen}, our generation process gets significantly accelerated by the off-the-shelf distillation technique~\cite{lightx2v}.
More importantly, as discussed in Sec.~\ref{sec:gen}, our bidirectional motion design enables more efficient reconstruction from sparse key frames without loss of generation performance.
\begin{table*}[ht]
\vspace{5pt}
\centering
\small
\setlength{\tabcolsep}{3pt}
\resizebox{0.95\textwidth}{!}{
\begin{tabular}{l|c|ccc|cccccc}
\toprule
\multirow{2}{*}{Method} & \multirow{2}{*}{Frames} & \multicolumn{3}{c|}{Inference Time (s)} & \multirow{2}{*}{Subj. Consist.} & \multirow{2}{*}{Back. Consist.} & \multirow{2}{*}{Temp. Flick.} & \multirow{2}{*}{Motion Smooth.} & \multirow{2}{*}{Aesth. Quality} & \multirow{2}{*}{Imag. Quality} \\
&&Recon. &Gen.&Total&&&&&&\\
\midrule
TrajectoryCrafter~\cite{trajectorycrafter}&49&25&121&146&83.02&88.58&94.71&97.64&44.63&54.59\\
ReCamMaster~\cite{recammaster}&81&-&168&168&88.21&91.60&96.56&\textbf{98.86}&44.29&58.87\\
\midrule
Ours (11 key frames)&81&2&18&\textbf{20}&88.43&92.27&\textbf{96.77}&98.80&44.55&59.75\\
Ours (21 key frames)&81&3&18&21&88.73&92.43&96.76&98.71&44.59&60.01\\
Ours (41 key frames)&81&5&18&23&89.10&92.65&96.67&98.63&\textbf{44.89}&60.37\\
Ours (full frames)&81&10&18&28&\textbf{89.42}&\textbf{92.79}&96.51&98.67&44.78&\textbf{61.51}\\
\bottomrule
\end{tabular}
}
\caption{\textbf{VBench~\cite{vbench} results for novel view generation.} We randomly collect 100 unseen in-the-wild videos, each with 4 different camera trajectories, resulting in a total of 400 test cases. For a fair comparison of inference time, we resize all videos to $336 \times 560$ resolution and report the average results over all test cases. The runtime evaluation is conducted on an A800 GPU.
}
\vspace{-13pt}
\label{tab:generation}
\end{table*}

\subsection{Qualitative Evaluation and Analysis}
\begin{figure}[h!]
    \centering
    \includegraphics[width=\linewidth]{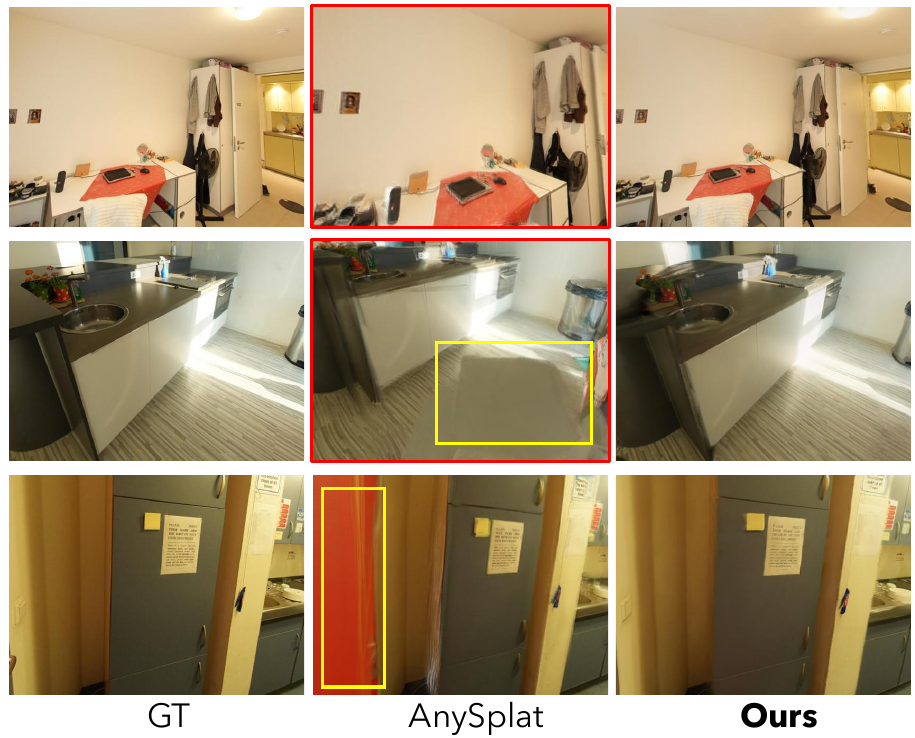}
    \vspace{-5mm}
    \caption{Qualitative comparison with state-of-the-art methods in \textbf{static scenes}. Red boundaries indicate inconsistent renderings due to inaccurate pose prediction. Yellow boxes indicate artifacts.}
    \label{fig:static_comp}
\end{figure}

\begin{figure}[ht]
    \centering
    \includegraphics[width=\linewidth]{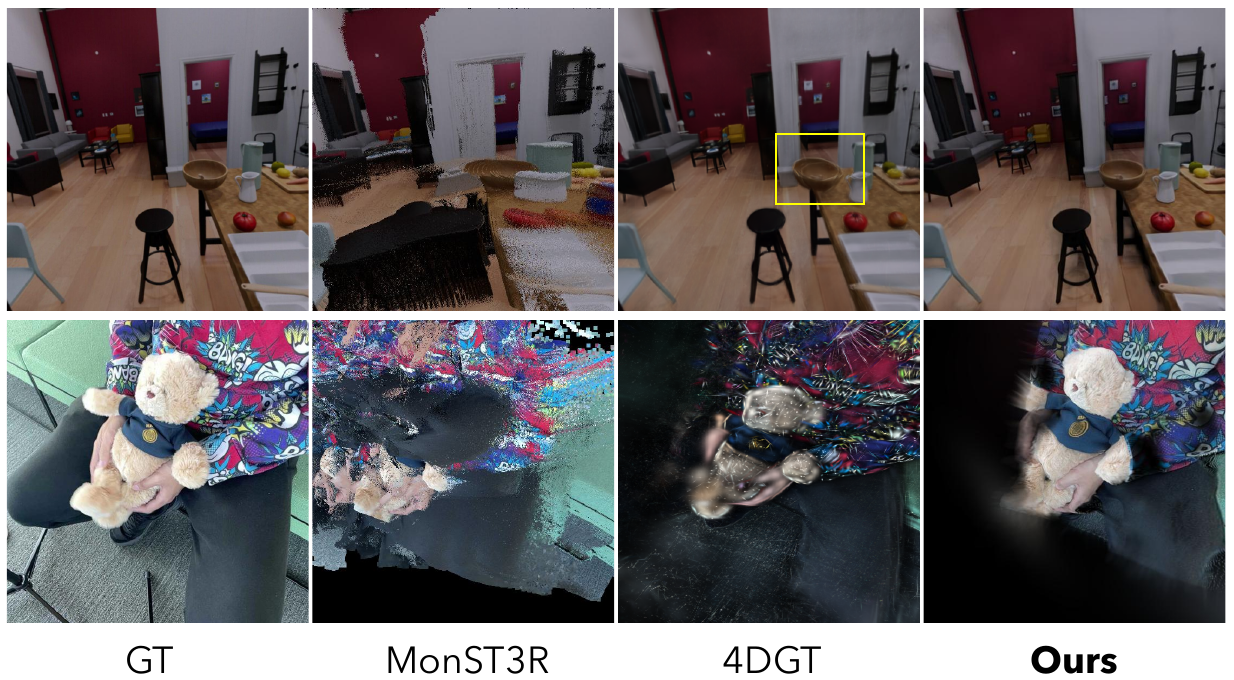}
    \vspace{-6mm}
    \caption{Qualitative comparison with state-of-the-art methods in \textbf{dynamic scenes}. Yellow boxes indicate artifacts.
    Note that the black regions in our prediction are \emph{not error} but mainly caused by partial observations of input frames.
    }
    \vspace{-3mm}
    \label{fig:dyna_comp}
\end{figure}

For an intuitive understanding, we conduct rich qualitative evaluations and analysis, leading to the following findings.
\vspace{-3mm}
\paragraph{Rendering quality.}
Fig.~\ref{fig:static_comp} and Fig.~\ref{fig:dyna_comp} demonstrate the rendering quality comparison. Our model not only achieves better visual quality but is also more faithful to input observations.
Instead, other methods may predict unreal artifacts such as regions indicated by yellow boxes in Fig.~\ref{fig:static_comp}.
\vspace{-3mm}
\paragraph{Pose prediction accuracy.}
It is noteworthy that our model also has better pose prediction accuracy. In Fig.~\ref{fig:static_comp}, the compared method~\cite{anysplat} shows a field of view (images with red boundaries) inconsistent with the ground truth, which is caused by inaccurate pose prediction.
\vspace{-3mm}
\paragraph{Trajectory controlability vs. generation quality.}
An intriguing and fundamental phenomenon we can find in Fig.~\ref{fig:gen_comp} is that related work usually demonstrates a trade-off between generation quality and trajectory controllability.
Specifically, TrajectoryCrater, a reconstruction-generation hybrid method similar to our \name, shows good trajectory controllability and exhibits consistent trajectories with our method, while its generation quality is inferior.
This is mainly caused by its non-scalable training pipeline, stopping the model from seeing diverse in-the-wild videos, such as very challenging human activities in Fig.~\ref{fig:gen_comp}.
\par
In contrast, the purely generation-based method ReCamMaster shows good visual generation quality, but cannot achieve precise trajectory control, which is crucial in some downstream tasks such as simulation.
\vspace{-3mm}
\paragraph{Artifact suppression.}
Another reason for our superiority over the similar reconstruction-based TrajectoryCrafter is that our degradation simulations (Fig.~\ref{fig:degradation}) enable artifact suppression.
In contrast, the generation quality of TrajectoryCrafter is significantly decreased by ``ghosting patterns'' from inaccurate reconstruction.
\vspace{-3mm}
\paragraph{Contextually grounded imagination.} Fig.~\ref{fig:gen_comp} also demonstrates that our \name can conduct contextually grounded imagination for non-observed regions, such as the second singer and crowded people.
We give credit to our design scalability to diverse in-the-wild videos.

\subsection{Ablation Study}
\vspace{-3mm}
\begin{table}[h]
\vspace{5pt}
\centering
\footnotesize
\resizebox{\columnwidth}{!}{
\begin{tabular}{l|ccc}
\toprule
Method& PSNR$\uparrow$ & SSIM$\uparrow$ & LPIPS$\downarrow$\\
\midrule
w/o Regularization&10.86&0.244&0.576\\
w/o Bidirectional Motion &11.27&0.285&0.570\\
Reconstruction part &11.56&0.293&0.558\\
w/ Generation& 14.59 & 0.323& 0.501\\
\bottomrule
\end{tabular}
}
\caption{\textbf{Ablation experiments on DyCheck.} ``w/. Generation'' indicates our full pipeline, which gains significant performance improvements over the pure reconstruction part.}
\vspace{-13pt}
\label{tab:ablation}
\end{table}

\begin{figure}[ht]
    \centering
    \includegraphics[width=\linewidth]{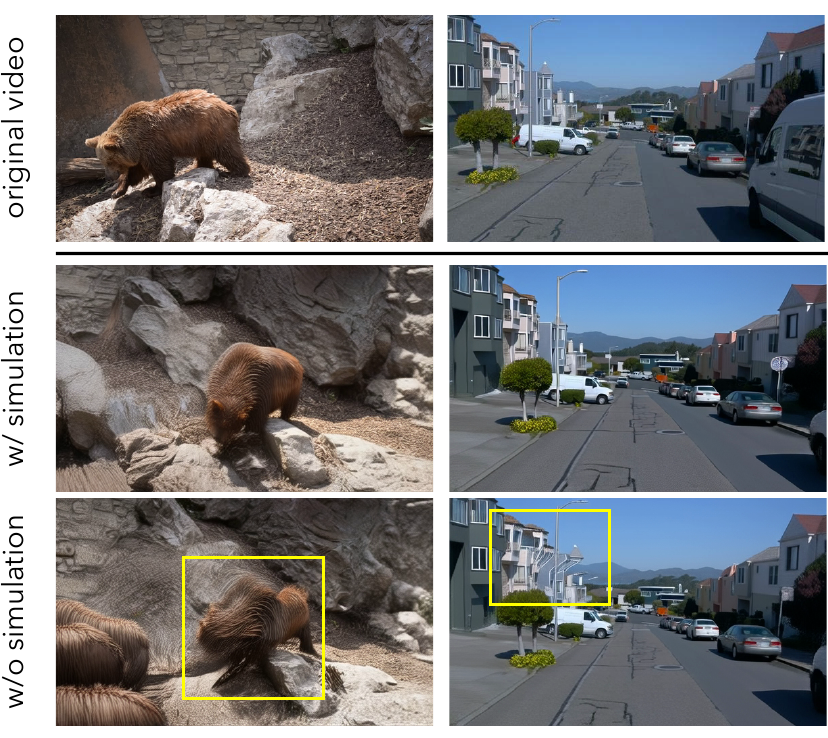}
    \vspace{-6mm}
    \caption{\textbf{Effectiveness of degradation simulation.} The model learns to suppress artifacts and hallucinate realistic details in occluded or distorted regions through degradation simulation.}
    \vspace{-4mm}
    \label{fig:ablation_degrad}
\end{figure}

\begin{figure}[ht]
    \centering
    \includegraphics[width=\linewidth]{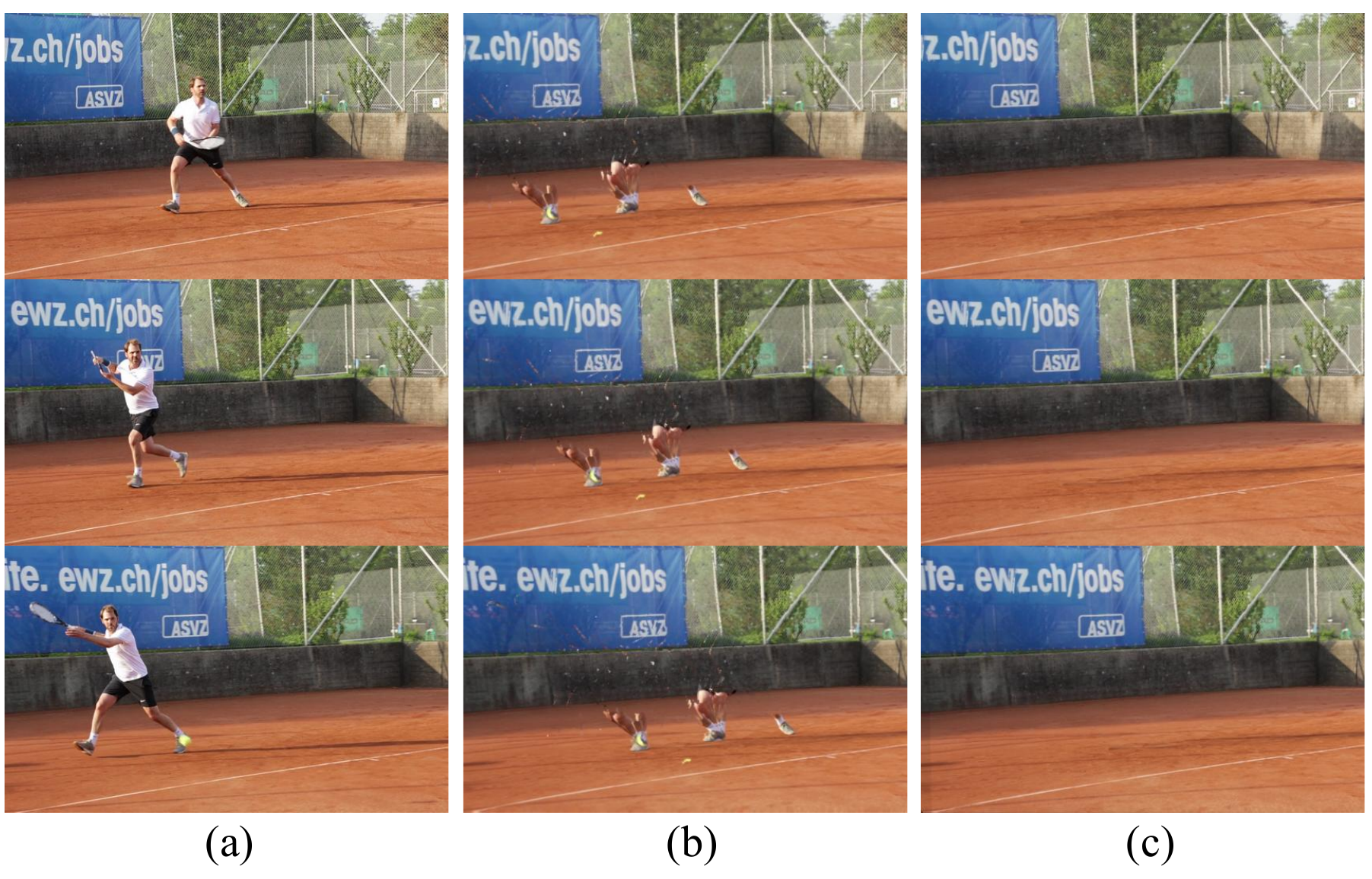}
    \vspace{-6mm}
    \caption{\textbf{Visualization about global motion tracking and aggregation}. (a) Input video. (b) Aggregated static Gaussians separated by predicted velocities. (c) Aggregated static Gaussians separated with global motion tracking.}
    \vspace{-3mm}
    \label{fig:background}
\end{figure}

\vspace{-3mm}
\paragraph{Motion modeling.}
In Table~\ref{tab:ablation}, we remove the motion modeling mechanism by skipping \cref{eq:motion} and predicting motions directly from frame features.
The performance drop reveals the effectiveness of our modeling mechanism.
\vspace{-3mm}
\paragraph{Opacity regularization.} In Sec.~\ref{sec:train}, we introduce opacity regularization to avoid the model learning a shortcut, which is outputting transparent primitives for the regions in similar colors to the predefined background color.
This technique is proven effective in Table~\ref{tab:ablation}.
\vspace{-3mm}
\paragraph{Degradation simulation.}
As discussed in \cref{sec:gen}, large camera motions often result in degraded renderings containing flying edge pixels and distortions.
\cref{fig:ablation_degrad} demonstrates the necessity of our online degradation simulation.
Without training on simulated degraded samples, the generation model tends to trust the geometric artifacts in the condition, leading to ``ghosting'' effects or blurred outputs.
\textbf{By incorporating degradation simulation, the model learns to suppress these artifacts and hallucinate realistic details in occluded or distorted regions}.
\paragraph{Global motion tracking.} Fig.~\ref{fig:background} showcases the importance of global motion tracking when identifying the dynamic instances.
Without the global tracking, some dynamic objects are mistakenly identified as static due to a partial static state.

\subsection{Applications}
\label{sec:application}
\begin{figure}[ht]
    \centering
    \vspace{-3mm}
    \includegraphics[width=0.9\linewidth]{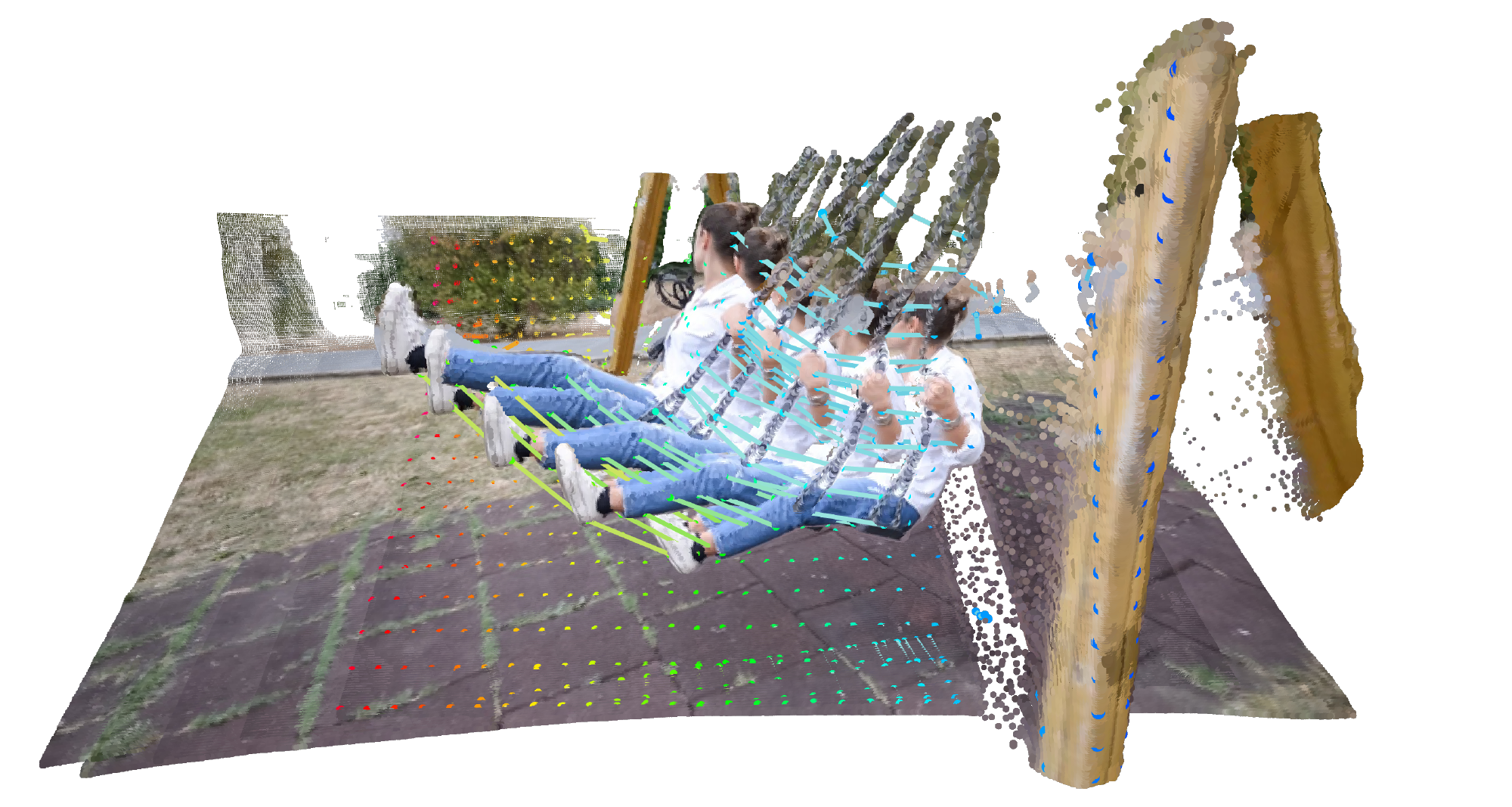}
    \vspace{-2mm}
    \caption{\textbf{Visualization of 3D tracking.} For better visualization, we only show the Gaussian centers.}
    \vspace{-6mm}
    \label{fig:flow}
\end{figure}

\begin{figure}[h]
    \centering
    \includegraphics[width=\linewidth]{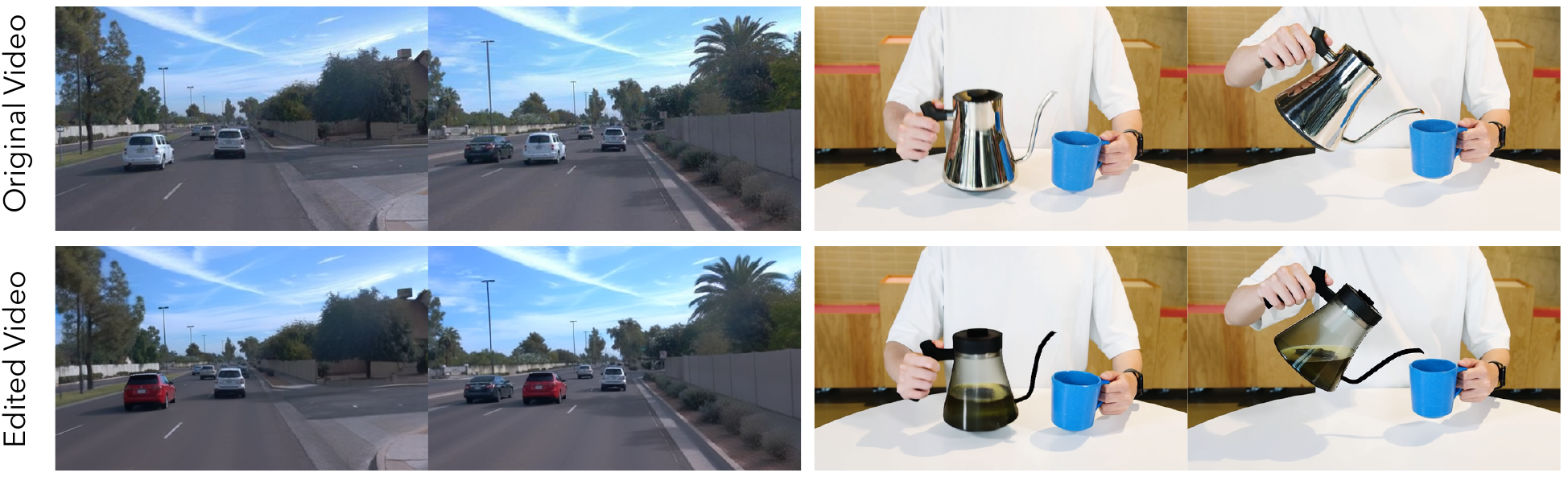}
    \caption{\textbf{Video editing.} Left: The white car is edited to be red. Right: The mirror teapot is edited to be transparent.}
    \vspace{-3mm}
    \label{fig:edition}
\end{figure}

A superiority of \name is the support for rich downstream applications other than the novel trajectory video generation.
Due to the limited space, here we briefly introduce several typical applications, leaving more details in the supplementary materials.
\vspace{-3mm}
\paragraph{3D tracking.}
By associating nearest Gaussian primitives between consecutive frames using predicted 3D flow, our \name achieves 3D tracking shown in Fig.~\ref{fig:flow}.
\vspace{-3mm}
\paragraph{Video editing.}
Since our model has a binary mask condition and a textual condition, it can edit videos with the help of a video segmentation model~\cite{sam2}, demonstrated in Fig.~\ref{fig:edition}.
\vspace{-3mm}
\paragraph{Video stabilization.}
By smoothing the predicted camera trajectory, our model achieves effective video stabilization, as demonstrated in the teaser Fig.~\ref{fig:teaser}.
\vspace{-5mm}
\paragraph{Video super-resolution}
The Gaussian representation in \name supports flexible rendering resolution without the significant loss of appearance information.
Thus, \name can achieve video super-resolution by generation with a larger rendering resolution, also demonstrated in Fig.~\ref{fig:teaser}.
\vspace{-3mm}
\paragraph{Others.} Moreover, \name is also capable of other applications such as background extraction (Fig.~\ref{fig:background}), image to world (Fig.~\ref{fig:teaser}).
We leave more demonstrations in the supplementary materials.

\section{Conclusion and Limitations}

In this paper, we introduce \textbf{\namenospace}, a 4D world model that overcomes key scalability limitations in previous arts, building a training pipeline scalable to in-the-wild monocular videos.
Thus, the generalization and versatility of \name are significantly enhanced by the diverse in-the-wild data, enabling various downstream applications.
Extensive experiments demonstrate state-of-the-art performance in both reconstruction and generation tasks.
\vspace{-4mm}
\paragraph{Limitations.}
\name requires data with correct underlying 3D information.
Therefore, it cannot be trivially applied to data without 3D information like 2D cartoons.
Due to the constraints of training resources, our curated dataset (1M clips) is not that large. We leave more data for future work.

\section*{Acknowledgements}
{\looseness=-1
This work was supported in part by the National Natural Science Foundation of China (No.~62320106010) and in part by Beijing Natural Science Foundation (No.~L257004, No.~L257015).
\par}
{
    \small
    \bibliographystyle{ieeenat_fullname}
    \bibliography{main}

@String(CVPR= {IEEE Conf. Comput. Vis. Pattern Recog.})

@String(ICCV= {Int. Conf. Comput. Vis.})

@String(ECCV= {Eur. Conf. Comput. Vis.})

@String(NIPS= {Adv. Neural Inform. Process. Syst.})

@String(TOG= {ACM Trans. Graph.})

@String(ICLR = {Int. Conf. Learn. Represent.})

@String(CVPR  = {CVPR})

@String(ICCV  = {ICCV})

@String(ECCV  = {ECCV})

@String(NIPS  = {NeurIPS})

@String(TOG   = {ACM TOG})

@String(ICLR  = {ICLR})

@article{3dgs,
  title={3D Gaussian Splatting for Real-Time Radiance Field Rendering},
  author={Kerbl, Bernhard and Kopanas, Georgios and Leimk{\"u}hler, Thomas and Drettakis, George},
  journal={TOG},
  volume={42},
  number={4},
  pages={139--1},
  year={2023}
}

@inproceedings{vggt,
  title={VGGT: Visual Geometry Grounded Transformer},
  author={Wang, Jianyuan and Chen, Minghao and Karaev, Nikita and Vedaldi, Andrea and Rupprecht, Christian and Novotny, David},
  booktitle={CVPR},
  pages={5294--5306},
  year={2025}
}

@article{4dgt,
  title={4DGT: Learning a 4D Gaussian Transformer Using Real-World Monocular Videos},
  author={Xu, Zhen and Li, Zhengqin and Dong, Zhao and Zhou, Xiaowei and Newcombe, Richard and Lv, Zhaoyang},
  journal={arXiv preprint arXiv:2506.08015},
  year={2025}
}

@inproceedings{noposplat,
  title={No Pose, No Problem: Surprisingly Simple 3D Gaussian Splats from Sparse Unposed Images},
  author={Ye, Botao and Liu, Sifei and Xu, Haofei and Li, Xueting and Pollefeys, Marc and Yang, Ming-Hsuan and Peng, Songyou},
  booktitle={ICLR},
  year={2024}
}

@article{anysplat,
  title={AnySplat: Feed-forward 3D Gaussian Splatting from Unconstrained Views},
  author={Jiang, Lihan and Mao, Yucheng and Xu, Linning and Lu, Tao and Ren, Kerui and Jin, Yichen and Xu, Xudong and Yu, Mulin and Pang, Jiangmiao and Zhao, Feng and others},
  journal={arXiv preprint arXiv:2505.23716},
  year={2025}
}

@article{movies,
  title={MoVieS: Motion-Aware 4D Dynamic View Synthesis in One Second},
  author={Lin, Chenguo and Lin, Yuchen and Pan, Panwang and Yu, Yifan and Yan, Honglei and Fragkiadaki, Katerina and Mu, Yadong},
  journal={arXiv preprint arXiv:2507.10065},
  year={2025}
}

@article{streamsplat,
  title={StreamSplat: Towards Online Dynamic 3D Reconstruction from Uncalibrated Video Streams},
  author={Wu, Zike and Yan, Qi and Yi, Xuanyu and Wang, Lele and Liao, Renjie},
  journal={arXiv preprint arXiv:2506.08862},
  year={2025}
}

@inproceedings{recammaster,
  title={ReCamMaster: Camera-Controlled Generative Rendering from a Single Video},
  author={Bai, Jianhong and Xia, Menghan and Fu, Xiao and Wang, Xintao and Mu, Lianrui and Cao, Jinwen and Liu, Zuozhu and Hu, Haoji and Bai, Xiang and Wan, Pengfei and others},
  booktitle={ICCV},
  year={2025}
}

@inproceedings{trajectorycrafter,
  title={TrajectoryCrafter: Redirecting Camera Trajectory for Monocular Videos via Diffusion Models},
  author={YU, Mark and Hu, Wenbo and Xing, Jinbo and Shan, Ying},
  booktitle={ICCV},
  year={2025}
}

@article{viewcrafter,
  title={ViewCrafter: Taming Video Diffusion Models for High-Fidelity Novel View Synthesis},
  author={Yu, Wangbo and Xing, Jinbo and Yuan, Li and Hu, Wenbo and Li, Xiaoyu and Huang, Zhipeng and Gao, Xiangjun and Wong, Tien-Tsin and Shan, Ying and Tian, Yonghong},
  journal={arXiv preprint arXiv:2409.02048},
  year={2024}
}

@inproceedings{dust3r,
  title={DUSt3R: Geometric 3D Vision Made Easy},
  author={Wang, Shuzhe and Leroy, Vincent and Cabon, Yohann and Chidlovskii, Boris and Revaud, Jerome},
  booktitle={CVPR},
  pages={20697--20709},
  year={2024}
}

@article{dinov2,
  title={DINOv2: Learning Robust Visual Features without Supervision},
  author={Oquab, Maxime and Darcet, Timoth{\'e}e and Moutakanni, Th{\'e}o and Vo, Huy and Szafraniec, Marc and Khalidov, Vasil and Fernandez, Pierre and Haziza, Daniel and Massa, Francisco and El-Nouby, Alaaeldin and others},
  journal={arXiv preprint arXiv:2304.07193},
  year={2023}
}

@inproceedings{dpt,
  title={Vision Transformers for Dense Prediction},
  author={Ranftl, Ren{\'e} and Bochkovskiy, Alexey and Koltun, Vladlen},
  booktitle={ICCV},
  pages={12179--12188},
  year={2021}
}

@article{wan,
  title={Wan: Open and Advanced Large-Scale Video Generative Models},
  author={Wan, Team and Wang, Ang and Ai, Baole and Wen, Bin and Mao, Chaojie and Xie, Chen-Wei and Chen, Di and Yu, Feiwu and Zhao, Haiming and Yang, Jianxiao and others},
  journal={arXiv preprint arXiv:2503.20314},
  year={2025}
}

@article{uni3c,
  title={Uni3C: Unifying Precisely 3D-Enhanced Camera and Human Motion Controls for Video Generation},
  author={Cao, Chenjie and Zhou, Jingkai and Li, Shikai and Liang, Jingyun and Yu, Chaohui and Wang, Fan and Xue, Xiangyang and Fu, Yanwei},
  journal={arXiv preprint arXiv:2504.14899},
  year={2025}
}

@article{voyager,
  title={Voyager: Long-Range and World-Consistent Video Diffusion for Explorable 3D Scene Generation},
  author={Huang, Tianyu and Zheng, Wangguandong and Wang, Tengfei and Liu, Yuhao and Wang, Zhenwei and Wu, Junta and Jiang, Jie and Li, Hui and Lau, Rynson WH and Zuo, Wangmeng and others},
  journal={arXiv preprint arXiv:2506.04225},
  year={2025}
}

@inproceedings{scannet++,
  title={ScanNet++: A High-Fidelity Dataset of 3D Indoor Scenes},
  author={Yeshwanth, Chandan and Liu, Yueh-Cheng and Nie{\ss}ner, Matthias and Dai, Angela},
  booktitle={ICCV},
  pages={12--22},
  year={2023}
}

@inproceedings{dl3dv,
  title={DL3DV-10K: A Large-Scale Scene Dataset for Deep Learning-Based 3D Vision},
  author={Ling, Lu and Sheng, Yichen and Tu, Zhi and Zhao, Wentian and Xin, Cheng and Wan, Kun and Yu, Lantao and Guo, Qianyu and Yu, Zixun and Lu, Yawen and others},
  booktitle={CVPR},
  pages={22160--22169},
  year={2024}
}

@inproceedings{waymo,
  title={Scalability in Perception for Autonomous Driving: Waymo Open Dataset},
  author={Sun, Pei and Kretzschmar, Henrik and Dotiwalla, Xerxes and Chouard, Aurelien and Patnaik, Vijaysai and Tsui, Paul and Guo, James and Zhou, Yin and Chai, Yuning and Caine, Benjamin and others},
  booktitle={CVPR},
  pages={2446--2454},
  year={2020}
}

@article{vkitti2,
  title={Virtual KITTI 2},
  author={Cabon, Yohann and Murray, Naila and Humenberger, Martin},
  journal={arXiv preprint arXiv:2001.10773},
  year={2020}
}

@inproceedings{DaS,
  title={Diffusion as Shader: 3D-Aware Video Diffusion for Versatile Video Generation Control},
  author={Gu, Zekai and Yan, Rui and Lu, Jiahao and Li, Peng and Dou, Zhiyang and Si, Chenyang and Dong, Zhen and Liu, Qifeng and Lin, Cheng and Liu, Ziwei and others},
  booktitle={Proceedings of the Special Interest Group on Computer Graphics and Interactive Techniques Conference Conference Papers},
  pages={1--12},
  year={2025}
}

@inproceedings{gen3c,
  title={GEN3C: 3D-Informed World-Consistent Video Generation with Precise Camera Control},
  author={Ren, Xuanchi and Shen, Tianchang and Huang, Jiahui and Ling, Huan and Lu, Yifan and Nimier-David, Merlin and M{\"u}ller, Thomas and Keller, Alexander and Fidler, Sanja and Gao, Jun},
  booktitle={CVPR},
  pages={6121--6132},
  year={2025}
}

@article{free4d,
  title={Free4D: Tuning-free 4D Scene Generation with Spatial-Temporal Consistency},
  author={Liu, Tianqi and Huang, Zihao and Chen, Zhaoxi and Wang, Guangcong and Hu, Shoukang and Shen, Liao and Sun, Huiqiang and Cao, Zhiguo and Li, Wei and Liu, Ziwei},
  journal={arXiv preprint arXiv:2503.20785},
  year={2025}
}

@inproceedings{freesim,
  title={FreeSim: Toward Free-Viewpoint Camera Simulation in Driving Scenes},
  author={Fan, Lue and Zhang, Hao and Wang, Qitai and Li, Hongsheng and Zhang, Zhaoxiang},
  booktitle={CVPR},
  pages={12004--12014},
  year={2025}
}

@inproceedings{depthcrafter,
  title={DepthCrafter: Generating Consistent Long Depth Sequences for Open-World Videos},
  author={Hu, Wenbo and Gao, Xiangjun and Li, Xiaoyu and Zhao, Sijie and Cun, Xiaodong and Zhang, Yong and Quan, Long and Shan, Ying},
  booktitle={CVPR},
  pages={2005--2015},
  year={2025}
}

@article{yang2024depth,
  title={Depth Any Video with Scalable Synthetic Data},
  author={Yang, Honghui and Huang, Di and Yin, Wei and Shen, Chunhua and Liu, Haifeng and He, Xiaofei and Lin, Binbin and Ouyang, Wanli and He, Tong},
  journal={arXiv preprint arXiv:2410.10815},
  year={2024}
}

@article{cogvideo,
  title={CogVideo: Large-Scale Pretraining for Text-to-Video Generation via Transformers},
  author={Hong, Wenyi and Ding, Ming and Zheng, Wendi and Liu, Xinghan and Tang, Jie},
  journal={arXiv preprint arXiv:2205.15868},
  year={2022}
}

@article{cogvideox,
  title={CogVideoX: Text-to-Video Diffusion Models with an Expert Transformer},
  author={Yang, Zhuoyi and Teng, Jiayan and Zheng, Wendi and Ding, Ming and Huang, Shiyu and Xu, Jiazheng and Yang, Yuanming and Hong, Wenyi and Zhang, Xiaohan and Feng, Guanyu and others},
  journal={arXiv preprint arXiv:2408.06072},
  year={2024}
}

@article{hunyuan_dit,
  title={Hunyuan-DiT: A Powerful Multi-Resolution Diffusion Transformer with Fine-Grained Chinese Understanding},
  author={Li, Zhimin and Zhang, Jianwei and Lin, Qin and Xiong, Jiangfeng and Long, Yanxin and Deng, Xinchi and Zhang, Yingfang and Liu, Xingchao and Huang, Minbin and Xiao, Zedong and others},
  journal={arXiv preprint arXiv:2405.08748},
  year={2024}
}

@inproceedings{stable_diffusion,
  title={High-Resolution Image Synthesis with Latent Diffusion Models},
  author={Rombach, Robin and Blattmann, Andreas and Lorenz, Dominik and Esser, Patrick and Ommer, Bj{\"o}rn},
  booktitle={CVPR},
  pages={10684--10695},
  year={2022}
}

@article{ex4d,
  title={EX-4D: EXtreme Viewpoint 4D Video Synthesis via Depth Watertight Mesh},
  author={Hu, Tao and Peng, Haoyang and Liu, Xiao and Ma, Yuewen},
  journal={arXiv preprint arXiv:2506.05554},
  year={2025}
}

@article{flexworld,
  title={FlexWorld: Progressively Expanding 3D Scenes for Flexible-View Synthesis},
  author={Chen, Luxi and Zhou, Zihan and Zhao, Min and Wang, Yikai and Zhang, Ge and Huang, Wenhao and Sun, Hao and Wen, Ji-Rong and Li, Chongxuan},
  journal={arXiv preprint arXiv:2503.13265},
  year={2025}
}

@article{monst3r,
  title={MonST3R: A Simple Approach for Estimating Geometry in the Presence of Motion},
  author={Zhang, Junyi and Herrmann, Charles and Hur, Junhwa and Jampani, Varun and Darrell, Trevor and Cole, Forrester and Sun, Deqing and Yang, Ming-Hsuan},
  journal={arXiv preprint arXiv:2410.03825},
  year={2024}
}

@article{bochkovskii2024depth,
  title={Depth Pro: Sharp Monocular Metric Depth in Less Than a Second},
  author={Bochkovskii, Aleksei and Delaunoy, Ama{\~A}{\c{G}}l and Germain, Hugo and Santos, Marcel and Zhou, Yichao and Richter, Stephan R and Koltun, Vladlen},
  journal={arXiv preprint arXiv:2410.02073},
  year={2024}
}

@inproceedings{vivid4d,
  title={Vivid4D: Improving 4D Reconstruction from Monocular Video by Video Inpainting},
  author={Huang, Jiaxin and Miao, Sheng and Yang, Bangbang and Ma, Yuewen and Liao, Yiyi},
  booktitle={ICCV},
  pages={12592--12604},
  year={2025}
}

@inproceedings{difix3d,
  title={Difix3D+: Improving 3D Reconstructions with Single-Step Diffusion Models},
  author={Wu, Jay Zhangjie and Zhang, Yuxuan and Turki, Haithem and Ren, Xuanchi and Gao, Jun and Shou, Mike Zheng and Fidler, Sanja and Gojcic, Zan and Ling, Huan},
  booktitle={CVPR},
  pages={26024--26035},
  year={2025}
}

@inproceedings{syncammaster,
  title={SynCamMaster: Synchronizing Multi-Camera Video Generation from Diverse Viewpoints},
  author={Bai, Jianhong and Xia, Menghan and Wang, Xintao and Yuan, Ziyang and Fu, Xiao and Liu, Zuozhu and Hu, Haoji and Wan, Pengfei and Zhang, Di},
  booktitle={ICLR},
  year={2025}
}

@inproceedings{cut3r,
  title={Continuous 3D Perception Model with Persistent State},
  author={Wang, Qianqian and Zhang, Yifei and Holynski, Aleksander and Efros, Alexei A and Kanazawa, Angjoo},
  booktitle={CVPR},
  pages={10510--10522},
  year={2025}
}

@inproceedings{flare,
  title={FLARE: Feed-Forward Geometry, Appearance and Camera Estimation from Uncalibrated Sparse Views},
  author={Zhang, Shangzhan and Wang, Jianyuan and Xu, Yinghao and Xue, Nan and Rupprecht, Christian and Zhou, Xiaowei and Shen, Yujun and Wetzstein, Gordon},
  booktitle={CVPR},
  pages={21936--21947},
  year={2025}
}

@article{dycheck,
  title={Monocular Dynamic View Synthesis: A Reality Check},
  author={Gao, Hang and Li, Ruilong and Tulsiani, Shubham and Russell, Bryan and Kanazawa, Angjoo},
  journal={NIPS},
  volume={35},
  pages={33768--33780},
  year={2022}
}

@inproceedings{vrnerf,
  title={VR-NeRF: High-Fidelity Virtualized Walkable Spaces},
  author={Xu, Linning and Agrawal, Vasu and Laney, William and Garcia, Tony and Bansal, Aayush and Kim, Changil and Rota Bul{\`o}, Samuel and Porzi, Lorenzo and Kontschieder, Peter and Bo{\v{z}}i{\v{c}}, Alja{\v{z}} and others},
  booktitle={SIGGRAPH Asia},
  pages={1--12},
  year={2023}
}

@inproceedings{adt,
  title={Aria Digital Twin: A New Benchmark Dataset for Egocentric 3D Machine Perception},
  author={Pan, Xiaqing and Charron, Nicholas and Yang, Yongqian and Peters, Scott and Whelan, Thomas and Kong, Chen and Parkhi, Omkar and Newcombe, Richard and Ren, Yuheng Carl},
  booktitle={ICCV},
  pages={20133--20143},
  year={2023}
}

@article{pixelperfect,
  title={Pixel-Perfect Depth with Semantics-Prompted Diffusion Transformers},
  author={Xu, Gangwei and Lin, Haotong and Luo, Hongcheng and Wang, Xianqi and Yao, Jingfeng and Zhu, Lianghui and Pu, Yuechuan and Chi, Cheng and Sun, Haiyang and Wang, Bing and others},
  journal={arXiv preprint arXiv:2510.07316},
  year={2025}
}

@inproceedings{see3d,
  title={You See It, You Got It: Learning 3D Creation on Pose-Free Videos at Scale},
  author={Ma, Baorui and Gao, Huachen and Deng, Haoge and Luo, Zhengxiong and Huang, Tiejun and Tang, Lulu and Wang, Xinlong},
  booktitle={CVPR},
  pages={2016--2029},
  year={2025}
}

@inproceedings{controlnet,
  title={Adding Conditional Control to Text-to-Image Diffusion Models},
  author={Zhang, Lvmin and Rao, Anyi and Agrawala, Maneesh},
  booktitle={ICCV},
  pages={3836--3847},
  year={2023}
}

@inproceedings{lpips,
  title={The Unreasonable Effectiveness of Deep Features as a Perceptual Metric},
  author={Zhang, Richard and Isola, Phillip and Efros, Alexei A and Shechtman, Eli and Wang, Oliver},
  booktitle={CVPR},
  pages={586--595},
  year={2018}
}

@article{gsplat,
  title={gsplat: An Open-Source Library for Gaussian Splatting},
  author={Ye, Vickie and Li, Ruilong and Kerr, Justin and Turkulainen, Matias and Yi, Brent and Pan, Zhuoyang and Seiskari, Otto and Ye, Jianbo and Hu, Jeffrey and Tancik, Matthew and others},
  journal={Journal of Machine Learning Research},
  volume={26},
  number={34},
  pages={1--17},
  year={2025}
}

@inproceedings{rectified,
  title={Scaling Rectified Flow Transformers for High-Resolution Image Synthesis},
  author={Esser, Patrick and Kulal, Sumith and Blattmann, Andreas and Entezari, Rahim and M{\"u}ller, Jonas and Saini, Harry and Levi, Yam and Lorenz, Dominik and Sauer, Axel and Boesel, Frederic and others},
  booktitle={Forty-first international conference on machine learning},
  year={2024}
}

@article{umT5,
  title={UniMax: Fairer and More Effective Language Sampling for Large-Scale Multilingual Pretraining},
  author={Chung, Hyung Won and Constant, Noah and Garcia, Xavier and Roberts, Adam and Tay, Yi and Narang, Sharan and Firat, Orhan},
  journal={arXiv preprint arXiv:2304.09151},
  year={2023}
}

@article{vace,
  title={VACE: All-in-One Video Creation and Editing},
  author={Jiang, Zeyinzi and Han, Zhen and Mao, Chaojie and Zhang, Jingfeng and Pan, Yulin and Liu, Yu},
  journal={arXiv preprint arXiv:2503.07598},
  year={2025}
}

@article{lora,
  title={LoRA: Low-Rank Adaptation of Large Language Models},
  author={Hu, Edward J and Shen, Yelong and Wallis, Phillip and Allen-Zhu, Zeyuan and Li, Yuanzhi and Wang, Shean and Wang, Lu and Chen, Weizhu and others},
  journal={ICLR},
  volume={1},
  number={2},
  pages={3},
  year={2022}
}

@misc{lightx2v,
 author = {LightX2V Contributors},
 title = {LightX2V: Light Video Generation Inference Framework},
 year = {2025},
 publisher = {GitHub},
 journal = {GitHub repository},
 howpublished = {\url{https://github.com/ModelTC/lightx2v}},
}

@inproceedings{pointodyssey,
  title={PointOdyssey: A Large-Scale Synthetic Dataset for Long-Term Point Tracking},
  author={Zheng, Yang and Harley, Adam W and Shen, Bokui and Wetzstein, Gordon and Guibas, Leonidas J},
  booktitle={ICCV},
  pages={19855--19865},
  year={2023}
}

@article{arkitscenes,
  title={ARKitScenes: A Diverse Real-World Dataset for 3D Indoor Scene Understanding Using Mobile RGB-D Data},
  author={Baruch, Gilad and Chen, Zhuoyuan and Dehghan, Afshin and Dimry, Tal and Feigin, Yuri and Fu, Peter and Gebauer, Thomas and Joffe, Brandon and Kurz, Daniel and Schwartz, Arik and others},
  journal={arXiv preprint arXiv:2111.08897},
  year={2021}
}

@inproceedings{dynamicstereo,
  title={DynamicStereo: Consistent Dynamic Depth from Stereo Videos},
  author={Karaev, Nikita and Rocco, Ignacio and Graham, Benjamin and Neverova, Natalia and Vedaldi, Andrea and Rupprecht, Christian},
  booktitle={CVPR},
  pages={13229--13239},
  year={2023}
}

@inproceedings{kubric,
  title={Kubric: A Scalable Dataset Generator},
  author={Greff, Klaus and Belletti, Francois and Beyer, Lucas and Doersch, Carl and Du, Yilun and Duckworth, Daniel and Fleet, David J and Gnanapragasam, Dan and Golemo, Florian and Herrmann, Charles and others},
  booktitle={CVPR},
  pages={3749--3761},
  year={2022}
}

@inproceedings{gfie,
  title={GFIE: A Dataset and Baseline for Gaze-Following from 2D to 3D in Indoor Environments},
  author={Hu, Zhengxi and Yang, Yuxue and Zhai, Xiaolin and Yang, Dingye and Zhou, Bohan and Liu, Jingtai},
  booktitle={CVPR},
  pages={8907--8916},
  year={2023}
}

@article{spatialvid,
  title={SpatialVid: A Large-Scale Video Dataset with Spatial Annotations},
  author={Wang, Jiahao and Yuan, Yufeng and Zheng, Rujie and Lin, Youtian and Gao, Jian and Chen, Lin-Zhuo and Bao, Yajie and Zhang, Yi and Zeng, Chang and Zhou, Yanxi and others},
  journal={arXiv preprint arXiv:2509.09676},
  year={2025}
}

@inproceedings{hoi4d,
  title={HOI4D: A 4D Egocentric Dataset for Category-Level Human-Object Interaction},
  author={Liu, Yunze and Liu, Yun and Jiang, Che and Lyu, Kangbo and Wan, Weikang and Shen, Hao and Liang, Boqiang and Fu, Zhoujie and Wang, He and Yi, Li},
  booktitle={CVPR},
  pages={21013--21022},
  year={2022}
}

@inproceedings{spring,
  title={Spring: A High-Resolution High-Detail Dataset and Benchmark for Scene Flow, Optical Flow and Stereo},
  author={Mehl, Lukas and Schmalfuss, Jenny and Jahedi, Azin and Nalivayko, Yaroslava and Bruhn, Andr{\'e}s},
  booktitle={CVPR},
  pages={4981--4991},
  year={2023}
}

@inproceedings{vbench,
  title={VBench: Comprehensive Benchmark Suite for Video Generative Models},
  author={Huang, Ziqi and He, Yinan and Yu, Jiashuo and Zhang, Fan and Si, Chenyang and Jiang, Yuming and Zhang, Yuanhan and Wu, Tianxing and Jin, Qingyang and Chanpaisit, Nattapol and others},
  booktitle={CVPR},
  pages={21807--21818},
  year={2024}
}

@article{sam2,
  title={SAM 2: Segment Anything in Images and Videos},
  author={Ravi, Nikhila and Gabeur, Valentin and Hu, Yuan-Ting and Hu, Ronghang and Ryali, Chaitanya and Ma, Tengyu and Khedr, Haitham and R{\"a}dle, Roman and Rolland, Chloe and Gustafson, Laura and others},
  journal={arXiv preprint arXiv:2408.00714},
  year={2024}
}

@inproceedings{tartanair,
  title={TartanAir: A Dataset to Push the Limits of Visual SLAM},
  author={Wang, Wenshan and Zhu, Delong and Wang, Xiangwei and Hu, Yaoyu and Qiu, Yuheng and Wang, Chen and Hu, Yafei and Kapoor, Ashish and Scherer, Sebastian},
  booktitle={IROS},
  pages={4909--4916},
  year={2020},
}

@inproceedings{bedlam,
  title={BEDLAM: A Synthetic Dataset of Bodies Exhibiting Detailed Lifelike Animated Motion},
  author={Black, Michael J and Patel, Priyanka and Tesch, Joachim and Yang, Jinlong},
  booktitle={CVPR},
  pages={8726--8737},
  year={2023}
}

@inproceedings{deepmvs,
  title={DeepMVS: Learning Multi-View Stereopsis},
  author={Huang, Po-Han and Matzen, Kevin and Kopf, Johannes and Ahuja, Narendra and Huang, Jia-Bin},
  booktitle={CVPR},
  pages={2821--2830},
  year={2018}
}

@inproceedings{cop3d,
  title={Common Pets in 3D: Dynamic New-View Synthesis of Real-Life Deformable Categories},
  author={Sinha, Samarth and Shapovalov, Roman and Reizenstein, Jeremy and Rocco, Ignacio and Neverova, Natalia and Vedaldi, Andrea and Novotny, David},
  booktitle={Proceedings of the IEEE/CVF Conference on Computer Vision and Pattern Recognition},
  pages={4881--4891},
  year={2023}
}

@inproceedings{hypersim,
  title={Hypersim: A Photorealistic Synthetic Dataset for Holistic Indoor Scene Understanding},
  author={Roberts, Mike and Ramapuram, Jason and Ranjan, Anurag and Kumar, Atulit and Bautista, Miguel Angel and Paczan, Nathan and Webb, Russ and Susskind, Joshua M},
  booktitle={ICCV},
  pages={10912--10922},
  year={2021}
}

@inproceedings{mapfree,
  title={Map-Free Visual Relocalization: Metric Pose Relative to a Single Image},
  author={Arnold, Eduardo and Wynn, Jamie and Vicente, Sara and Garcia-Hernando, Guillermo and Monszpart, Aron and Prisacariu, Victor and Turmukhambetov, Daniyar and Brachmann, Eric},
  booktitle={ECCV},
  pages={690--708},
  year={2022},
  organization={Springer}
}

@article{worldmirror,
  title={WorldMirror: Universal 3D World Reconstruction with Any-Prior Prompting},
  author={Liu, Yifan and Min, Zhiyuan and Wang, Zhenwei and Wu, Junta and Wang, Tengfei and Yuan, Yixuan and Luo, Yawei and Guo, Chunchao},
  journal={arXiv preprint arXiv:2510.10726},
  year={2025}
}

@article{mapanything,
  title={MapAnything: Universal Feed-Forward Metric 3D Reconstruction},
  author={Keetha, Nikhil and M{\"u}ller, Norman and Sch{\"o}nberger, Johannes and Porzi, Lorenzo and Zhang, Yuchen and Fischer, Tobias and Knapitsch, Arno and Zauss, Duncan and Weber, Ethan and Antunes, Nelson and others},
  journal={arXiv preprint arXiv:2509.13414},
  year={2025}
}

@article{da3,
  title={Depth Anything 3: Recovering the Visual Space from Any Views},
  author={Lin, Haotong and Chen, Sili and Liew, Junhao and Chen, Donny Y and Li, Zhenyu and Shi, Guang and Feng, Jiashi and Kang, Bingyi},
  journal={arXiv preprint arXiv:2511.10647},
  year={2025}
}

@article{worldforge,
  title={WorldForge: Unlocking Emergent 3D/4D Generation in Video Diffusion Model via Training-Free Guidance},
  author={Song, Chenxi and Yang, Yanming and Zhao, Tong and Li, Ruibo and Zhang, Chi},
  journal={arXiv preprint arXiv:2509.15130},
  year={2025}
}

@article{see4d,
  title={SEE4D: Pose-Free 4D Generation via Auto-Regressive Video Inpainting},
  author={Lu, Dongyue and Liang, Ao and Huang, Tianxin and Fu, Xiao and Zhao, Yuyang and Ma, Baorui and Pan, Liang and Yin, Wei and Kong, Lingdong and Ooi, Wei Tsang and others},
  journal={arXiv preprint arXiv:2510.26796},
  year={2025}
}

@article{postcam,
  title={PostCam: Camera-Controllable Novel-View Video Generation with Query-Shared Cross-Attention},
  author={Chen, Yipeng and Ye, Zhichao and Fang, Zhenzhou and Chen, Xinyu and Zhang, Xiaoyu and Liu, Jialing and Wang, Nan and Liu, Haomin and Zhang, Guofeng},
  journal={arXiv preprint arXiv:2511.17185},
  year={2025}
}

@article{chronosobserver,
  title={ChronosObserver: Taming 4D World with Hyperspace Diffusion Sampling},
  author={Wang, Qisen and Zhao, Yifan and Shen, Peisen and Li, Jialu and Li, Jia},
  journal={arXiv preprint arXiv:2512.01481},
  year={2025}
}

@article{ma2025follow,
  title={Follow-Your-Creation: Empowering 4D Creation through Video Inpainting},
  author={Ma, Yue and Feng, Kunyu and Zhang, Xinhua and Liu, Hongyu and Zhang, David Junhao and Xing, Jinbo and Zhang, Yinhan and Yang, Ayden and Wang, Zeyu and Chen, Qifeng},
  journal={arXiv preprint arXiv:2506.04590},
  year={2025}
}

@article{lightx,
  title={Light-X: Generative 4D Video Rendering with Camera and Illumination Control},
  author={Liu, Tianqi and Chen, Zhaoxi and Huang, Zihao and Xu, Shaocong and Zhang, Saining and Ye, Chongjie and Li, Bohan and Cao, Zhiguo and Li, Wei and Zhao, Hao and others},
  journal={arXiv preprint arXiv:2512.05115},
  year={2025}
}

@InProceedings{layeranimate,
    author    = {Yang, Yuxue and Fan, Lue and Lin, Zuzeng and Wang, Feng and Zhang, Zhaoxiang},
    title     = {LayerAnimate: Layer-Level Control for Animation},
    booktitle = {ICCV},
    month     = {October},
    year      = {2025},
    pages     = {10865-10874}
}

@article{pvg,
  title={Periodic Vibration Gaussian: Dynamic Urban Scene Reconstruction and Real-Time Rendering},
  author={Chen, Yurui and Gu, Chun and Jiang, Junzhe and Zhu, Xiatian and Zhang, Li},
  journal={arXiv preprint arXiv:2311.18561},
  year={2023}
}

@article{omnire,
  title={OmniRe: Omni Urban Scene Reconstruction},
  author={Chen, Ziyu and Yang, Jiawei and Huang, Jiahui and de Lutio, Riccardo and Esturo, Janick Martinez and Ivanovic, Boris and Litany, Or and Gojcic, Zan and Fidler, Sanja and Pavone, Marco and others},
  journal={arXiv preprint arXiv:2408.16760},
  year={2024}
}

@inproceedings{centerpoint,
  title={Center-Based 3D Object Detection and Tracking},
  author={Yin, Tianwei and Zhou, Xingyi and Krahenbuhl, Philipp},
  booktitle={Proceedings of the IEEE/CVF conference on computer vision and pattern recognition},
  pages={11784--11793},
  year={2021}
}

@article{bevformer,
  title={BEVFormer: Learning Bird's-Eye-View Representation from LiDAR-Camera via Spatiotemporal Transformers},
  author={Li, Zhiqi and Wang, Wenhai and Li, Hongyang and Xie, Enze and Sima, Chonghao and Lu, Tong and Yu, Qiao and Dai, Jifeng},
  journal={IEEE Transactions on Pattern Analysis and Machine Intelligence},
  year={2024},
  publisher={IEEE}
}

@article{mixsup,
  title={MixSup: Mixed-Grained Supervision for Label-Efficient LiDAR-Based 3D Object Detection},
  author={Yang, Yuxue and Fan, Lue and Zhang, Zhaoxiang},
  journal={arXiv preprint arXiv:2401.16305},
  year={2024}
}

@inproceedings{sst,
  title={{Embracing Single Stride 3D Object Detector with Sparse Transformer}},
  author={Fan, Lue and Pang, Ziqi and Zhang, Tianyuan and Wang, Yu-Xiong and Zhao, Hang and Wang, Feng and Wang, Naiyan and Zhang, Zhaoxiang},
  booktitle={CVPR},
  year={2022}
}

@inproceedings{van2024generative,
  title={Generative Camera Dolly: Extreme Monocular Dynamic Novel View Synthesis},
  author={Van Hoorick, Basile and Wu, Rundi and Ozguroglu, Ege and Sargent, Kyle and Liu, Ruoshi and Tokmakov, Pavel and Dave, Achal and Zheng, Changxi and Vondrick, Carl},
  booktitle={European Conference on Computer Vision},
  pages={313--331},
  year={2024},
  organization={Springer}
}

@article{gsdit,
  title={GS-DiT: Advancing Video Generation with Pseudo 4D Gaussian Fields through Efficient Dense 3D Point Tracking},
  author={Bian, Weikang and Huang, Zhaoyang and Shi, Xiaoyu and Li, Yijin and Wang, Fu-Yun and Li, Hongsheng},
  journal={arXiv preprint arXiv:2501.02690},
  year={2025}
}

@inproceedings{camclonemaster,
  title={CamCloneMaster: Enabling Reference-based Camera Control for Video Generation},
  author={Luo, Yawen and Bai, Jianhong and Shi, Xiaoyu and Xia, Menghan and Wang, Xintao and Wan, Pengfei and Zhang, Di and Gai, Kun and Xue, Tianfan},
  booktitle={SIGGRAPH Asia},
  year={2025}
}

@inproceedings{cinemaster,
  title={CineMaster: A 3D-Aware and Controllable Framework for Cinematic Text-to-Video Generation},
  author={Wang, Qinghe and Luo, Yawen and Shi, Xiaoyu and Jia, Xu and Lu, Huchuan and Xue, Tianfan and Wang, Xintao and Wan, Pengfei and Zhang, Di and Gai, Kun},
  booktitle={SIGGRAPH},
  pages={1--10},
  year={2025}
}

@inproceedings{flexdrive,
  title={FlexDrive: Toward Trajectory Flexibility in Driving Scene Gaussian Splatting Reconstruction and Rendering},
  author={Zhou, Jingqiu and Fan, Lue and Huang, Linjiang and Shi, Xiaoyu and Liu, Si and Zhang, Zhaoxiang and Li, Hongsheng},
  booktitle={CVPR},
  pages={1549--1558},
  year={2025}
}

@inproceedings{five,
  title={Five-bench: A fine-grained video editing benchmark for evaluating emerging diffusion and rectified flow models},
  author={Li, Minghan and Xie, Chenxi and Wu, Yichen and Zhang, Lei and Wang, Mengyu},
  booktitle={ICCV},
  pages={16672--16681},
  year={2025}
}

@article{anyv2v,
  title={Anyv2v: A tuning-free framework for any video-to-video editing tasks},
  author={Ku, Max and Wei, Cong and Ren, Weiming and Yang, Harry and Chen, Wenhu},
  journal={arXiv preprint arXiv:2403.14468},
  year={2024}
}

@article{tapvid3d,
  title={Tapvid-3d: A benchmark for tracking any point in 3d},
  author={Koppula, Skanda and Rocco, Ignacio and Yang, Yi and Heyward, Joe and Carreira, Joao and Zisserman, Andrew and Brostow, Gabriel and Doersch, Carl},
  journal={NIPS},
  volume={37},
  pages={82149--82165},
  year={2024}
}

@inproceedings{spatialtracker,
  title={Spatialtracker: Tracking any 2d pixels in 3d space},
  author={Xiao, Yuxi and Wang, Qianqian and Zhang, Shangzhan and Xue, Nan and Peng, Sida and Shen, Yujun and Zhou, Xiaowei},
  booktitle={CVPR},
  pages={20406--20417},
  year={2024}
}

@inproceedings{st4rtrack,
  title={St4rtrack: Simultaneous 4d reconstruction and tracking in the world},
  author={Feng, Haiwen and Zhang, Junyi and Wang, Qianqian and Ye, Yufei and Yu, Pengcheng and Black, Michael J and Darrell, Trevor and Kanazawa, Angjoo},
  booktitle={ICCV},
  pages={8503--8513},
  year={2025}
}
}

\clearpage
\setcounter{table}{0}
\setcounter{figure}{0}
\setcounter{equation}{0}
\setcounter{footnote}{0}
\renewcommand{\thetable}{S\arabic{table}}
\renewcommand{\thefigure}{S\arabic{figure}}
\renewcommand{\theequation}{S\arabic{equation}}
\maketitlesupplementary
\appendix

We provide \textcolor{blue}{videos on the project page}\footnote{\url{https://neoverse-4d.github.io}} to vividly present qualitative results for an enhanced view experience.

\section{Implementation Details}
\paragraph{Reconstruction model.}
The transformer decoders in the bidirectional motion-encoding branch follow the design of DUSt3R~\cite{dust3r}, where each decoder block consists of a self-attention layer for intra-frame spatial modeling and a cross-attention layer for inter-frame temporal modeling.
Finally, two DPT~\cite{dpt} heads are employed to predict the forward and backward motions, respectively.
Here, we define the forward/backward velocities $\{\boldsymbol{v}_i^+, \boldsymbol{v}_i^-\}$ as the 3D displacements from the current frame to the next/previous frame in the camera coordinate.
\paragraph{Generation model.}
The multiple encoders for multi-modal conditions are implemented with 1) VAE~\cite{wan} encoder for RGB images and depth maps, 2) convolutional layers with $8\times$ spatial and $4\times$ temporal compression ratio for masks and plüker embeddings.
During the generation training stage, only convolutional layers are trainable while the VAE encoder is frozen.
\section{Training Details}
To ensure compatibility with the patch size of DINOv2~\cite{dinov2} in the reconstruction model ($\times 14$ downsampling) and the VAE in the generation model ($\times 8$ compression), we resize all input videos to have a longest edge of 560 pixels during reconstruction training, and a fixed resolution of $336\times 560$ during generation training.
\paragraph{Reconstruction model.}
We train the reconstruction model on a combination of static and dynamic 3D datasets.
For each training iteration, we sample $N$ key frames (where $2\leq N \leq 8$) and $N-1$ intermediate target frames between adjacent key frames.
While only the $N$ key frames are processed by the reconstruction model to predict Gaussians, the supervision loss is computed on all $2N-1$ frames.
We utilize a cosine learning rate schedule with a peak learning rate of $1\times 10^{-4}$ and a warmup 5K iterations.
To enhance the model's robustness to temporal direction, we apply a random temporal reversal augmentation with a probability of $0.5$.
The weights for the multi-task loss (Eq. 6 in the main paper) are set as follows: $\lambda_{1}=5.0$ (camera), $\lambda_{2}=1.0$ (depth), $\lambda_{3}=1.0$ (motion), and $\lambda_{4}=0.1$ (regularization).
\vspace{-2mm}
\paragraph{Generation model.}
For the generation model, we use a constant learning rate of $1\times 10^{-5}$ and a batch size of 1 per GPU.
To enable efficient on-the-fly reconstruction, we randomly sample $11 \sim 21$ keyframes from each video clip to reconstruct the 4DGS representation.
Additionally, we employ a mask drop strategy where we randomly set all masks to 0 (indicating all degraded renderings need inpainting) with a probability of $0.2$ to improve model robustness.
\begin{table}[t]
\vspace{5pt}
\centering
\footnotesize
\resizebox{\columnwidth}{!}{
\begin{tabular}{@{}c@{\hspace{0.4em}}l|ccccc|c}
\toprule
& Dataset& Dynamic & Depth & Pose & Flow & Real & Clip \\
\midrule
\multirow{6}{*}{\large\ding{172}}&PointOdyssey~\cite{pointodyssey} &\ding{51}&\ding{51}&\ding{51}&\ding{51}& &131\\
&DynamicReplica~\cite{dynamicstereo} &\ding{51}&\ding{51}&\ding{51}&\ding{51}& &483\\
&Kubric~\cite{kubric} &\ding{51}&\ding{51}&\ding{51}&\ding{51}& &5.7K\\
&Spring~\cite{spring} &\ding{51}&\ding{51}&\ding{51}&\ding{51}& &37\\
&VKITTI2~\cite{vkitti2} &\ding{51}&\ding{51}&\ding{51}&\ding{51}& &50\\
&Waymo~\cite{waymo} &\ding{51}&\ding{51}&\ding{51}&\ding{51}&\ding{51}&798\\
\midrule
\multirow{4}{*}{\large\ding{173}}&TartanAir~\cite{tartanair} &\ding{51}&\ding{51}&\ding{51}& & &369\\
&BEDLAM~\cite{bedlam} &\ding{51}&\ding{51}&\ding{51}& & &10.4K\\
&MVS-Synth~\cite{deepmvs} &\ding{51}&\ding{51}&\ding{51}& & &120\\
&GFIE~\cite{gfie} &\ding{51}&\ding{51}&\ding{51}& &\ding{51}&81\\
\midrule
\multirow{2}{*}{\large\ding{174}}&HOI4D~\cite{hoi4d} &\ding{51}&\ding{51}& & &\ding{51}&3.0K\\
&CoP3D~\cite{cop3d} &\ding{51}& &\ding{51}& &\ding{51}&2.8K\\
\midrule
\multirow{5}{*}{\large\ding{175}}&DL3DV~\cite{dl3dv} & &\ding{51}&\ding{51}&\ding{51}&\ding{51}&6.4K\\
&Scannet++~\cite{scannet++} & &\ding{51}&\ding{51}&\ding{51}&\ding{51}&853\\
&ARKitScenes~\cite{arkitscenes} & &\ding{51}&\ding{51}&\ding{51}&\ding{51}&4.5K\\
&HyperSim~\cite{hypersim} & &\ding{51}&\ding{51}&\ding{51}&&457\\
&MapFree~\cite{mapfree} & &\ding{51}&\ding{51}&\ding{51}&\ding{51}&460\\
\midrule
\multirow{2}{*}{\large\ding{176}}&SpatialVID$^\dagger$~\cite{spatialvid} &\ding{51}&\ding{51}&\ding{51}& &\ding{51}&371.3K\\
&\textbf{Monocular Videos}&\ding{51}& & & &\ding{51}&\textbf{1M}\\
\bottomrule
\end{tabular}
}
\vspace{-2mm}
\caption{\textbf{Training Datasets.} We categorize existing datasets into 5 groups based on their data characteristics. Group \normalsize\ding{172}$\sim$\normalsize\ding{175} are used in reconstruction training, while group \normalsize\ding{176} is used in generation training.
$^\dagger$: we only use videos for generation training.}
\vspace{-5mm}
\label{tab:train_dataset}
\end{table}

\vspace{-6mm}
\section{Dataset Details}
We summarize the datasets used in our training in Table~\ref{tab:train_dataset}.
Our training data is categorized into five groups:
\begin{enumerate}
    \item [\ding{172}] Dynamic datasets with 3D flow for velocity supervision.
    \item [\ding{173}] Dynamic datasets with depth and camera poses.
    \item [\ding{174}] Dynamic datasets with incomplete 3D information (e.g., only camera poses or depth).
    \item [\ding{175}] Static datasets (we assume 3D flow is zero).
    \item [\ding{176}] Monocular videos.
\end{enumerate}
We train the reconstruction model on \ding{172} to \ding{175}, while the generation model is trained on \ding{176}.
Though SpatialVID provides 3D information, we don't use it for reconstruction training due to its unstable depth quality.

\section{Evaluation Protocol}
Following AnySplat, we perform test-time pose alignment to facilitate fair comparison, without introducing ground-truth poses during inference.
\paragraph{Static reconstruction.}
We evaluate static reconstruction performance on VRNeRF~\cite{vrnerf} and Scannet++~\cite{scannet++}.
\begin{itemize}
    \item \textbf{VRNeRF:} We select 6 scenes captured with pinhole cameras. For each scene, we randomly sample 16 views as input for reconstruction and 8 novel views for testing.
    \item \textbf{Scannet++:} We evaluate on all 50 scenes in the test set. We utilize 32 input views for reconstruction and evaluate on 16 novel views.
\end{itemize}
\paragraph{Dynamic reconstruction.}
For dynamic reconstruction on ADT~\cite{adt}, we follow 4DGT~\cite{4dgt} to evaluate the same 4 scenes:
\begin{itemize}
    \item Apartment\_release\_multiuser\_cook\_seq141\_M1292
    \item Apartment\_release\_multiskeleton\_party\_seq114\_M1292
    \item Apartment\_release\_meal\_skeleton\_seq135\_M1292
    \item Apartment\_release\_work\_skeleton\_seq137\_M1292
\end{itemize}
For each sequence, we sample a clip of 64 consecutive frames. We use 32 frames (stride 2) as input and the remaining 32 interleaved frames for testing.

For DyCheck~\cite{dycheck}, we evaluate 5 scenes (apple, block, paper-windmill, spin, teddy).
We sample 64 consecutive timestamps for each scene, using 32 frames (stride 2) from a casually-captured video (camera 0) for reconstruction and the complete 64 frames from another fixed-camera video (camera 1) for testing.

\begin{table}[t]
    \centering
    \footnotesize
    \resizebox{\linewidth}{!}{
    \begin{tabular}{l|cccc}
    \toprule
     & Struct.\ Dist.$\downarrow$ & CLIP Score $\uparrow$ & NIQE$\downarrow$ & Second Per Frame$\downarrow$\\
    \midrule
    AnyV2V~\cite{anyv2v}&0.071 &24.89 &5.04 & 6.11\\
    Wan-Edit~\cite{five}&\textbf{0.013} &26.39 &6.54 & 3.07\\
    VACE~\cite{vace}&0.015 &\textbf{26.92} &\textbf{4.37} & 4.30\\
    Ours&0.018 &26.66 &5.13 & \textbf{0.49} \\
    \bottomrule
    \end{tabular}
    }
    \caption{\textbf{Video editing evaluation on FiVE~\cite{five}.}}
    \label{tab:editing}
\end{table}
\begin{table}[t]
    \centering
    \footnotesize
    \resizebox{\linewidth}{!}{
    \begin{tabular}{l|ccc}
    \toprule
     & SpatialTracker~\cite{spatialtracker} & St4RTrack~\cite{st4rtrack} & Ours\\
    \midrule
    APD ($\delta_{3D}=0.1$m)$\uparrow$&3.79 &2.47 &\textbf{7.31}\\
    EPE$\downarrow$&3.35 &5.64 &\textbf{3.10}\\
    \bottomrule
    \end{tabular}
    }
    \caption{\textbf{3D tracking evaluation on DriveTrack through TAPVid-3D~\cite{tapvid3d}.} The prediction of SpatialTracker is offered by TAPVid-3D and St4RTrack is predicted from its official codebase.}
    \label{tab:tracking}
\end{table}

\section{Downstream Task Evaluation}
In the main paper, we qualitatively demonstrate several downstream applications of \namenospace.
Here, we provide quantitative evaluations on two representative tasks: video editing and 3D tracking.

\paragraph{Video editing.}
We evaluate video editing on the FiVE~\cite{five} benchmark and compare with AnyV2V~\cite{anyv2v}, Wan-Edit~\cite{five}, and VACE~\cite{vace}.
As shown in \cref{tab:editing}, although \name is not specifically designed for video editing, it achieves competitive performance while being significantly faster (0.49 seconds per frame vs.\ 3.07--6.11 seconds per frame for other methods).

\paragraph{3D tracking.}
We evaluate 3D tracking on the DriveTrack subset of TAPVid-3D~\cite{tapvid3d} and compare with SpatialTracker~\cite{spatialtracker} and St4RTrack~\cite{st4rtrack}.
As shown in \cref{tab:tracking}, it demonstrates that the 3D flow predicted by our reconstruction model provides reliable 3D correspondences.

\section{Discussion on Linear Motion Assumption}
As described in \cref{eq:mu}, our method assumes approximately linear motion between adjacent key frames for Gaussian interpolation.
While this is a simplified assumption, it does not negatively affect the generation quality for the following reasons.
During training, we reconstruct from sparse key frames and render to all frames.
The less-accurate non-keyframe renderings naturally serve as a form of temporal degradation, encouraging the generation model to learn to produce high-quality videos with non-linear motions from degraded renderings.
During inference, users can input all frames to ensure reliable renderings when needed.
Moreover, linear motion is a common and reasonable assumption adopted by prior works such as 4DGT~\cite{4dgt}, as real-world motion within short intervals between adjacent frames is generally well-approximated by linear interpolation.

\section{Limitations and Failure Cases}
\begin{figure}[t]
    \centering
    \includegraphics[width=\linewidth]{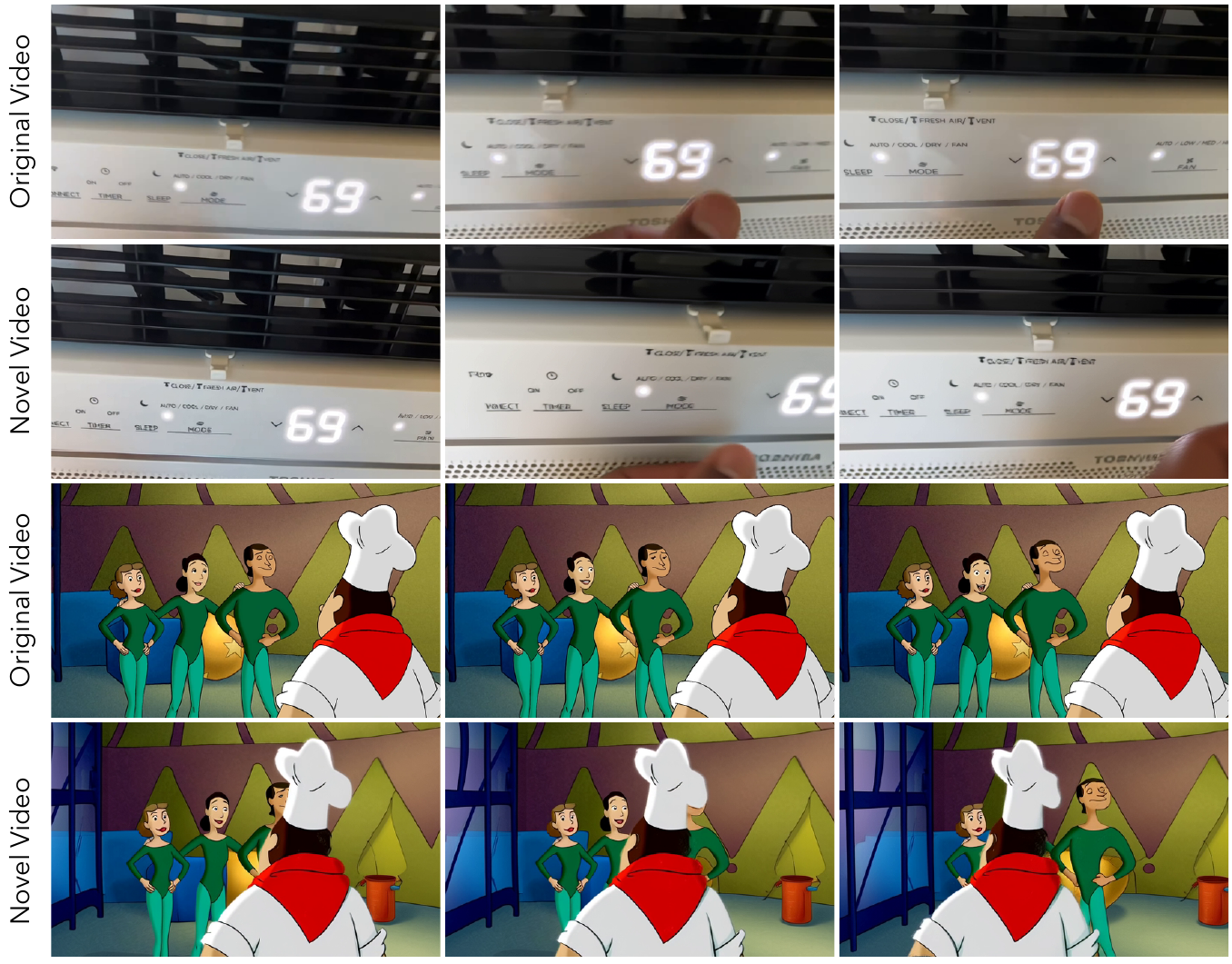}
    \caption{\textbf{Failure cases.} Top: Text generation failure. Bottom: Novel view generation on 2D data.}
    \label{fig:failure}
\end{figure}
Although our method can handle various challenging scenarios, there are some limitations as shown in \cref{fig:failure}.
Similar to many video diffusion models, our method occasionally \textbf{\textit{struggles to render legible and correct text}} (Top two rows).
Besides, our method relies on extracting 3D clues from videos.
It \textbf{\textit{struggles with data lacking 3D geometry, such as 2D cartoons}}.
For instance, as the camera moves to the right side of a 2D cartoon character (Bottom two rows), the model may fail to generate the correct 3D profile (e.g., revealing the other side of a face), as the input video lacks inherent 3D structure.

\section{Additional Qualitative Results}
\begin{figure}[t]
    \centering
    \vspace{-3mm}
    \includegraphics[width=\linewidth]{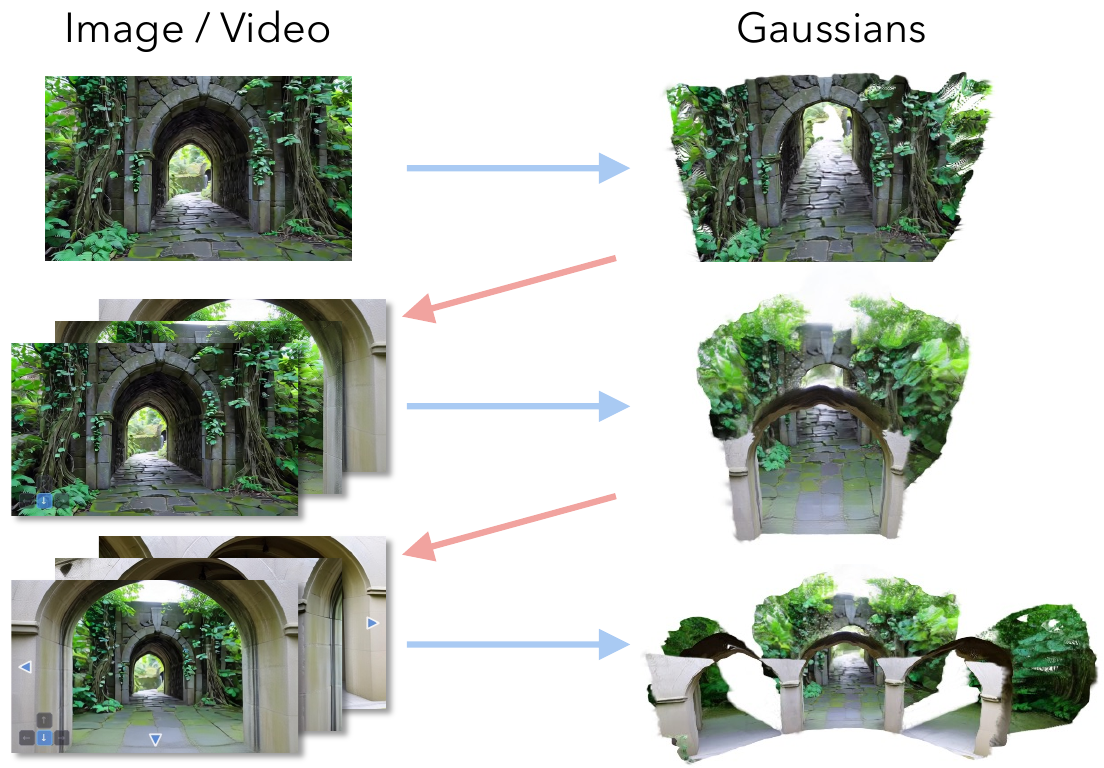}
    \caption{\textbf{Image to world.} Starting from a single view, \name can reconstruct a 3D scene, generate an exploration video, and iteratively expand the visible area.}
    \vspace{-3mm}
    \label{fig:env_expand}
\end{figure}
\paragraph{Image to world.}
Our \name allows for exploration in a captured image by iteratively generating new views and reconstructing the scene.
As illustrated in \cref{fig:env_expand}, given a single starting image, we can generate a spatially coherent video trajectory.
This generated video is then used to reconstruct a larger Gaussian Splatting scene, effectively "out-painting" the 3D world.

\begin{figure}[t]
    \centering
    \includegraphics[width=\linewidth]{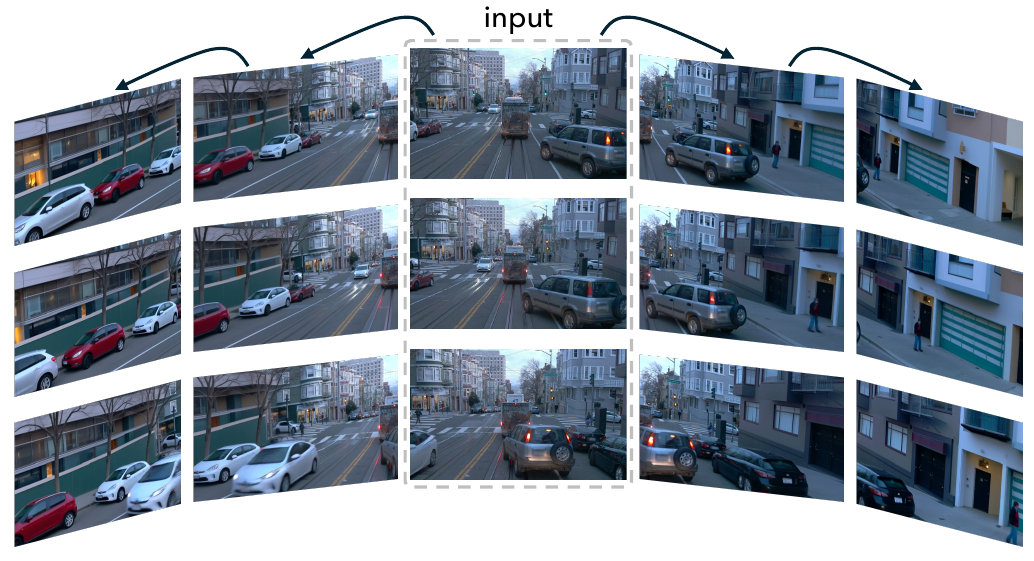}
    \caption{\textbf{Single-view to multi-view generation.} Starting from a single front-view video, \name can generate multi-view consistent videos.}
    \vspace{-5mm}
    \label{fig:driving_multiview}
\end{figure}
\paragraph{Single-view to multi-view}
\cref{fig:driving_multiview} demonstrates the capability of generating multi-view consistent videos from a single-view video through \textbf{iterative application of \namenospace}.

\end{document}